\documentclass[11pt]{amsart}

\usepackage[T1]{fontenc}
\usepackage{amsaddr}
\usepackage[sort,compress]{cite}
\usepackage{amsmath,amssymb,amsfonts}
\usepackage{multirow}
\usepackage{makecell}
\usepackage{colortbl}
\usepackage{siunitx}
\usepackage{booktabs}
\usepackage[sc]{mathpazo}
\usepackage{graphicx}
\usepackage{rotating}
\usepackage{paralist}
\usepackage{lscape}
\usepackage{nicefrac}
\usepackage{geometry}
\usepackage[table]{xcolor}
\usepackage{tikz,pgfplots}
\usepackage{enumitem}
\usepackage{mathtools}
\usepackage[toc,page]{appendix}
\usepackage{hyperref}

\usepackage{listings}

\newcolumntype{R}{>{\columncolor{gray!40}}r}
\newcolumntype{L}{>{\columncolor{gray!40}}l}
\newcolumntype{C}{>{\columncolor{gray!40}}c}

\newcommand{\figref}[1]{Figure~\ref{#1}}
\newcommand{\tabref}[1]{Table~\ref{#1}}
\newcommand{\secref}[1]{\S\ref{#1}}

\newcommand{\appref}[1]{Appendix~\ref{#1}}

\newcommand{\claire}{\texttt{CLAIRE}}
\newcommand{\defeq}{\ensuremath{\mathrel{\mathop:}=}}

\DeclareMathOperator*{\minopt}{minimize}

\newcommand{\ns}[1]{\ensuremath{\mathbb{#1}}}
\newcommand{\fun}[1]{\ensuremath{\mathcal{#1}}}
\newcommand{\vect}[1]{\boldsymbol{#1}}
\newcommand{\mat}[1]{\boldsymbol{#1}}
\newcommand{\idiv}{\ensuremath{\nabla\cdot}}
\newcommand{\igrad}{\ensuremath{\nabla}}

\newcommand{\dop}[1]{\ensuremath{\mathcal{#1}}}
\newcommand{\acr}[1]{\textbf{#1}}
\newcommand{\half}[1]{\frac{#1}{2}}
\newcommand{\dt}{\d{t}}
\newcommand{\dx}{\mathrm{d}\vect{x}}
\newcommand{\p}{\partial}

\definecolor{light-gray}{gray}{0.80}

\renewcommand\paragraph{\subsubsection*}

\title[CLAIRE---Parallelized Diffeomorphic Image Registration]{CLAIRE---Parallelized Diffeomorphic Image Registration for Large-Scale Biomedical Imaging Applications}

\author{Naveen Himthani$^{1}$, Malte Brunn$^{2}$, Jae-Youn Kim$^{3}$, Miriam Schulte$^{2}$, Andreas Mang$^{3}$
\and George Biros$^{1}$}
\address{${}^1$~Oden Institute, The University of Texas at Austin, Austin, TX, USA \\
${}^2$~Institute for Parallel and Distributed Systems, University of Stuttgart, Stuttgart, DE\\
${}^3$~Department of Mathematics, University of Houston, Houston, TX, USA}

\begin{document}
\maketitle

\begin{abstract}
We study the performance of \claire{}---a diffeomorphic multi-node, multi-GPU image registration algorithm and software---in large-scale biomedical imaging applications with billions of voxels. At such resolutions, most existing software packages for diffeomorphic image registration are prohibitively expensive. As a result, practitioners first significantly downsample the original images and then register them using existing tools. Our main contribution is an extensive analysis of the impact of downsampling on registration performance. We study this impact by comparing full resolution registrations obtained with \claire{} to lower resolution registrations for synthetic and real-world imaging datasets. Our results suggest that registration at full resolution can yield a superior registration quality---but not always. For example, downsampling a synthetic image from $1024^3$ to $256^3$ decreases the Dice coefficient from 92\% to 79\%. However, the differences are less pronounced for noisy or low contrast high resolution images. \claire{} allows us not only to register images of clinically relevant size in a few seconds but also to register images at unprecedented resolution in reasonable time. The highest resolution considered are CLARITY images of size $2816 \times 3016 \times 1162$. To the best of our knowledge, this is the first study on image registration quality at such resolutions.
\end{abstract}

\smallskip{\noindent\small\textbf{Keywords.} large-scale biomedical image processing, diffeomorphic image registration, high-\\performance computing, GPUs}

\section{Introduction}\label{s:intro}

\begin{figure}
\centering
\includegraphics[width=\textwidth]{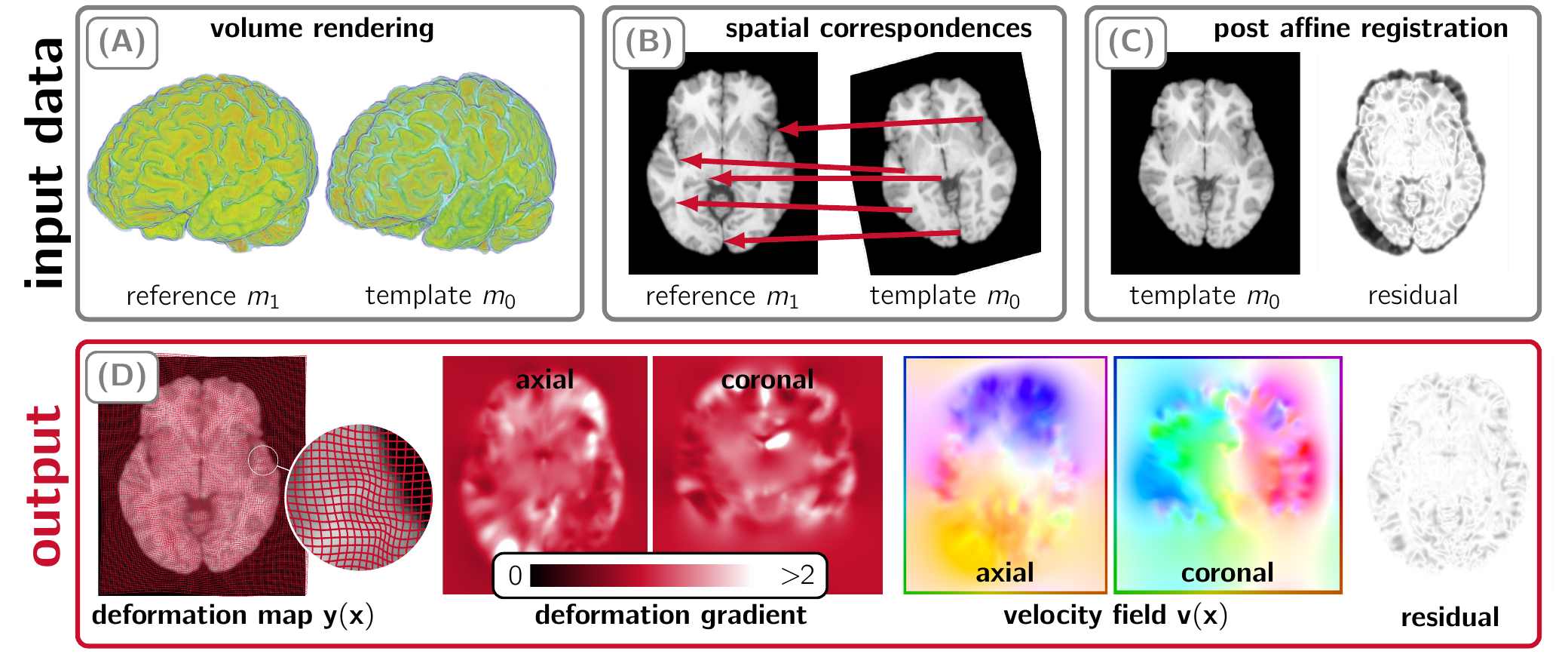}
\caption{Illustration of the image registration problem. Panel (A): 3D rendering of an exemplary set of input images. Panel (B): Image registration is the task of computing spatial correspondences between two images of the same object. These correspondences, denoted as $\vect{y}$, are indicated by the red arrows for example points within a single axial slice of the template and reference image data shown in panel (A). Before we execute \claire{}, we compensate for the global mismatch between the considered images by performing an affine registration. In panel (C), we show an axial slice of the volume shown in panel (A) after an affine pre-registration step has been carried out; we execute \claire{} on these images. Panel (D): \claire{} outputs a diffeomorphic deformation map $\vect{y}$ that matches each point in the template image $m_0$ to its corresponding point in the reference image $m_1$. We show a typical deformation map $\vect{y}$ in the leftmost image and the corresponding determinant of the deformation gradient (encodes volume change) in the second and third image from the left (axial and coronal slice). In \claire{}, we invert for a stationary velocity field $\vect{v}$ that parameterizes $\vect{y}$ (second and third figure from the right; color denotes orientation). The last figure in panel (D) shows the point-wise residual after applying \claire{}.}\label{fig:imregprob}
\end{figure}

3D diffeomorphic image registration (also known as ``image alignment'' or ``matching'') is a critical task in biomedical image analysis~\cite{Hajnal:2001a,Sotiras:2013a}. For example, it enables the study of morphological changes associated with the progression of neurodegenerative diseases over time or in imaging studies of patient populations. The process of image registration involves finding a spatial transformation which maps corresponding points in an image to those in another~\cite{Hajnal:2001a}. In mathematical notation, we are given two images $m_0(\vect{x})$ (the template/moving image) and $m_1(\vect{x})$ (the reference/fixed image; here $\vect{x}\in\Omega\subset\mathbb{R}^3)$ and we seek a spatial transformation $\vect{y} : \mathbb{R}^{3} \to \mathbb{R}^{3}$, such that the deformed template image $m_0(\vect{y}(\vect{x}))$ is similar to the reference image $m_1(\vect{x})$ for all $\vect{x}$ (see~\figref{fig:imregprob} for an illustration)~\cite{Modersitzki:2004a,Modersitzki:2008a}. Image registration methods can be categorized based on the parameterization for $\vect{y}$~\cite{Modersitzki:2004a}. We seek a diffeomorphic map $\vect{y}$, i.e., $\vect{y}$ is a differentiable bijection and has a differentiable inverse. Methods that parameterize $\vect{y}$ in terms of a smooth, time-varying velocity field $\vect{v} : \mathbb{R}^3 \times [0,1] \to\mathbb{R}^3$ belong to a class of methods referred to as \emph{large-deformation diffeomorphic metric mapping} (\acr{LDDMM})~\cite{Beg:2005a,Trouve:1998a,Younes:2010a}. In this study, we consider a related class of methods that use stationary velocity fields $\vect{v} : \mathbb{R}^3 \to\mathbb{R}^3$. This diffeomorphic registration problem is expensive to solve because the problem is infinite-dimensional, and upon discretization results in a nonlinear system with millions of unknowns---even for stationary velocity fields. For example, solving the registration problem for two images of resolution $256^3$ (a typical size for clinical scans) requires solving for approximately 50 million unknowns in space (three vector components per image grid point). Furthermore, image registration is a highly nonlinear, ill-posed inverse problem~\cite{Fischer:2008a}, resulting in ill-conditioned inversion operators. Consequently, running registration on multi-core high-end CPUs can take several minutes.

There exist various algorithms and software packages for fast registration of images at standard clinical resolution (e.g., $256^3$)~\cite{Vercauteren:2009a,Vercauteren:2007a,Avants:2014a,Yang:2017a,Krebs:2018a,Gu:2009a,Zhang:2019a,Grzech:2019a}. This includes \claire{}, which can execute image registration in parallel on multi-node multi-core CPUs and GPUs~\cite{Mang:2016SC,Gholami:2016IP,Mang:2018CLAIRE,Brunn:2020a,Brunn2021,Brunn:2020b}. We note that there is little work on scalable image registration. One application that requires this scalability is the registration of CLARITY images~\cite{Chung:2013b,Kim:2013a,Kutten:2016a,Kutten:2017a,Tomer:2014a,Vogelstein:2018a} with a resolution in the order of $20\text{\,K}\times20\text{\,K}\times1\text{\,K}$. This corresponds to a problem with approximately 1.2 trillion unknowns. In~\cite{Brunn:2020b}, we extended \claire{} to support GPU-accelerated scalable image registration, which can process high resolution images using multiple GPUs. We demonstrated the scalability of our solver using synthetic images with a resolution up to $2048^3$ and CLARITY mouse brain images of size $768\times768\times1024$. In this work, we scale registration to an even higher resolution, e.g., CLARITY images of size $2816\times3016\times1162$. This corresponds to an increase of 16$\times$ in problem size. In our previous work~\cite{Mang:2015NK,Mang:2016H1,Mang:2016SC,Mang:2017SL,Mang:2018CLAIRE,Mang:2018a,Gholami:2017SC,Mang:2017a}, we have extensively studied the algorithmic side of image registration within the framework of \claire{}. In this paper, we pay closer attention to the quality of the registration results. We study the effect of different input parameters, including the quality and resolution of the input images, on the accuracy of the registration.

\subsection{Contributions}

We build up on our prior work on scalable deformable image registration~\cite{Mang:2015NK,Mang:2016H1,Mang:2017SL,Mang:2018CLAIRE,Mang:2016SC,Mang:2018a,Mang:2017a,Brunn:2020a,Brunn:2020b,Brunn2021} using \claire{} and analyze the effect of image resolution on image registration accuracy. The present work analyzes image registration performance. We do not propose any major improvements in our methodology, with the exception of additional advice for hyperparameter tuning. We outline our past contributions on formulations, algorithms, and their parallel implementation below. Our major contributions of the present work are:
\begin{itemize}
\item We evaluate \claire{} on high resolution synthetic and real imaging datasets. We demonstrate that image registration when performed at native high resolution results in higher accuracy (measured in terms of the Dice coefficient of the labeled structured in the images). We conduct experiments to show that downsampling the images and then registering them result in loss of registration accuracy.
\item We design scalable image registration experiments to explore the effect of solver parameters---the number of time steps $n_t$ in the semi-Lagrangian scheme, and regularization parameters $\beta_v$ and $\beta_w$---on the registration performance.
\item We present an extension of the regularization parameter continuation scheme first presented in~\cite{Mang:2015NK} by searching for $\beta_w$ in addition to $\beta_v$, thereby removing the need for selecting an additional resolution-dependent solver parameter.
\item We study the performance of our scalable registration solver \claire{} for applications in high resolution mouse and human neuroimage registration. We perform image registration for two pairs of CLARITY mouse brain images at a resolution of $2816\times3016\times1162$ voxels. To the best of our knowledge, images of this scale have not been registered before at full resolution in under 30\,min.
\end{itemize}

\subsection{Related Work}

The current work builds upon the open source framework \claire{}~\cite{Mang:2015NK,Mang:2016H1,Mang:2016SC,Mang:2017SL,Mang:2018CLAIRE,Mang:2018a,Brunn:2020a,Brunn:2020b,Brunn2021,Gholami:2017SC,claire-web}. Our formulation for diffeomorphic image registration has been described in~\cite{Mang:2015NK,Mang:2016H1}. Our Newton--Krylov solver was originally developed in~\cite{Mang:2015NK}. We proposed efficient numerical implementations for evaluating forward and adjoint operators in~\cite{Mang:2017SL,Mang:2017a,Mang:2016SC}. We designed various methods for preconditioning in~\cite{Mang:2015NK,Mang:2017SL,Brunn:2020b,Mang:2018CLAIRE}. The computational kernels of the parallel CPU implementation of our solver were introduced in~\cite{Gholami:2017SC,Mang:2016SC,Mang:2018CLAIRE}. More recently, we ported \claire{} to GPU architectures~\cite{Brunn:2020a,Brunn:2020b}. In summary, our work paved the way towards real-time applications of diffeomorphic image registration and its deployment to high-resolution medical imaging application. To the best of our knowledge, this is the only existing software for diffeomorphic image registration with these capabilities. We have integrated our framework with biophysical modeling in~\cite{Gholami:2017SC,Mang:2018a,Mang:2020a,Scheufele:2019a,Scheufele:2020a}. None of these works explores registration performance in large-scale biomedical imaging applications.

Literature surveys of image registration and associated algorithmic developments can be found in~\cite{Modersitzki:2004a,Sotiras:2013a}. A recent overview of existing LDDMM methods can be found in~\cite{Mang:2018CLAIRE}. Related LDDMM software packages include \texttt{Demons} \cite{Vercauteren:2009a}, \texttt{ANTs} \cite{Avants:2011a,Avants:2008a,ants-web}, \texttt{DARTEL} \cite{Ashburner:2007a}, \texttt{deformetrica}~\cite{Bone:2018a,Bone:2018b,Fishbaugh:2017a,deformetrica-web}, \texttt{FLASH} \cite{Zhang:2018a}, \texttt{LDDMM} \cite{Beg:2005a,lddmm-web}, \texttt{ARDENT} \cite{ardent-web}, \texttt{ITKNDReg} \cite{itkndreg-web}, and \texttt{PyCA}~\cite{pyca-web}. Surveys of GPU-accelerated image registration solvers can be found in~\cite{Fluck:2011a,Shams:2010a,Eklund:2013a}; particular examples for various formulations are~\cite{Budelmann:2019a,Bone:2018b,Courty:2008a,Durrleman:2014a,Ellingwood:2016a,Gu:2009a,Grzech:2019a,Ha:2009a,Ha:2011a,Joshi:2005a,Koenig:2018a,Modat:2010a,Sommer:2011a,Shackleford:2010a,Shamonin:2014a,ValeroLara:2013a,ValeroLara:2014a}. Multi-GPU LDDMM implementations for atlas construction are described in~\cite{Ha:2009a,Ha:2011a,ValeroLara:2013a,ValeroLara:2014a}. Their setup is embarrassingly parallel in the sense that they solve many small registration problems independently on single GPUs. In~\cite{Ha:2009a,Ha:2011a,Joshi:2005a}, the computation bottlenecks are the repeated solution of a Helmholtz-type PDE and trilinear scattered data interpolation to compute and apply the deformation map. They use hardware acceleration for the trilinear interpolation kernel with 3D texture volume support. The runtime for a single dataset of size $160\times192\times160$ is \SI{20}{\second} on an NVIDIA Quadro FX 5600. \claire{} uses a multi-node multi-GPU framework with high computational throughput for single (large-scale) registration problems~\cite{Brunn:2020b} which is no longer an embarrassingly parallel problem.  \claire{} uses the Message Passing Interface (\acr{MPI}) to parallelize the implementation.

None of the GPU-accelerated LDDMM methods mentioned above, except for \claire{}~\cite{Mang:2015NK,Mang:2016H1,Mang:2016SC,Mang:2017SL,Mang:2018CLAIRE,Mang:2018a,Brunn:2020a,Brunn:2020b,Brunn2021,Gholami:2017SC,claire-web}, use second-order numerical optimization. Many of the available methods solve the registration problem by reducing the number of unknowns either through a coarse parameterization or by using a coarse grid and use simplified algorithms. These crude approximations and simplifications can result in inferior registration quality~\cite{Mang:2018CLAIRE,Brunn:2020a}.

The work in~\cite{Kutten:2016a} focuses on annotating CLARITY brain images by registering them to the Allen Institute's Mouse Reference Atlas ({\bf ARA}). They use a ``masked'' LDDMM approach. They also consider the registration of CLARITY-to-CLARITY brain images and compare different mismatch terms for the registrations. However, they downscale the images to a lower resolution for conducting all experiments. In~\cite{Kutten:2017a}, mutual information is used for the registration of CLARITY to the ARA dataset but at an approximately one hundred times downsampled resolution (at an original in-plane isotropic resolution 0.58~$\mu m$). The authors in~\cite{Nazib:2018a} analyze registration performance on high-resolution mouse brain images of size $2560\times2160\times633$ were obtained using the CUBIC protocol~\cite{Susaki:2015a}. They report results using different software packages including \texttt{ANTs} and \texttt{Elastix}. They did not observe a relationship between registration accuracy at different resolutions. For their high-resolution runs using \texttt{ANTs}, they report a wall clock time of over 200\,hrs on a single compute node (2.66GHz 64bit Intel Xeon processor with 256GB RAM) while the same run with \texttt{elastix}~\cite{Klein:2010a} took approximately $30$hrs. The authors in~\cite{Niedworok:2016a} register high-resolution images of mouse brains to the ARA dataset~\cite{Kuan:2015a}. They perform nonlinear registration using \texttt{ANTs} at coarse resolution ($10\mu m$ for the ARA) and apply the deformation at high-resolution. In the current work, we do not downsample high-resolution images but register them at the original resolution. We can register CLARITY images of resolution $2816\times3016\times1162$ in less than $30$min using 256 GPUs. In addition to that, we study the effect of resolution on the registration quality.

\subsection{Outline} We summarize the overall formulation in~\secref{s:formulation} and the algorithms in~\secref{s:disc_num_algo} for completeness. We note, that all of the material presented in~\secref{s:formulation} and~\secref{s:disc_num_algo} has been discussed in detail in~\cite{Brunn:2020b,Mang:2018CLAIRE}. In~\secref{sec:kernels}, we present our kernels and parallel algorithms and discuss key solver parameters. We also introduce a new scheme to automatically identify adequate parameters of our solver for unseen data. This scheme extends on our prior work in \cite{Mang:2018CLAIRE}. We conclude with the main scalability experiments in~\secref{s:results}, and present conclusions in~\secref{s:conclusions}.

\subsection{Limitations} \claire{} currently only supports mono-modal similarity measures, which limits our study to registrations for images acquired with the same imaging modality. Moreover, \claire{} only supports periodic boundary conditions, i.e., we require that the image data be embedded in a larger background domain. In most medical imaging applications, the images are embedded in a zero background and, therefore, naturally periodic. If the images are not periodic, they can be zero-padded and mollified. \claire{} uses stationary velocities, which improves computational efficiency, but it suboptimal from a theoretical point of view. In~\cite{Mang:2015NK}, we found no qualitative differences in registration mismatch when registering two images using stationary velocities. This observation is in line with the work of other groups using stationary velocity fields~\cite{Hernandez:2009a,Lorenzi:2013a}. Regarding computational performance, one issue is the memory requirement of our method. We have optimized memory allocation for the core components of \claire{}. Additional optimizations by reusing and sharing memory across external libraries to further reduce the memory load remain subject to future work.

\section{Methods}\label{s:methods}

Before discussing our enhancements in \secref{sec:kernels}, we shortly introduce the underlying mathematical formulation of the image registration problem utilized in \claire{} as well as the discretization and the numerical algorithms. The following exposition is included for completeness, only, and is based on material described in our prior work on efficient algorithms for diffeomorphic image registration~\cite{Mang:2015NK,Mang:2016H1,Mang:2016SC,Mang:2017SL,Mang:2017a,Mang:2018CLAIRE,Brunn:2020a,Brunn:2020b,Gholami:2017SC}. Consequently, we keep this section brief.

\begin{table}

\caption{Notation and main symbols.}\label{t:notation-and-symbols}
\centering\scriptsize
\begin{tabular}[t]{lllll}\toprule
\bf Symbol  & \bf  Description\\\midrule
$\Omega$ & spatial domain; $\Omega\defeq[0,2\pi)^3\subset\ns{R}^3$ with boundary $\p\Omega$ \\
$\vect{x}$ & spatial coordinate; $\vect{x}\defeq(x_1,x_2,x_3)\in\ns{R}^3$ \\
$t$ & (pseudo-)time variable; $t \in [0,1]$ \\
$m_1(\vect{x})$ & reference image (fixed image) \\
$m_0(\vect{x})$ & template image (moving image) \\
$\vect{v}(\vect{x})$ & stationary velocity field \\
$\vect{y}(\vect{x})$ & (diffeomorphic) deformation map \\
$m(\vect{x},t)$ & state variable (transported intensities of $m_0$) \\
$\lambda(\vect{x},t)$ & adjoint variable \\
$\dop{A}$ & regularization operator \\
$\beta_v > 0$ & regularization parameter for $\vect{v}$ \\
$\beta_w > 0$ & regularization parameter for $\nabla \cdot \vect{v}$ \\
$\mat{F}$ & deformation gradient \\
$J$ & determinant of deformation gradient (Jacobian determinant) \\
$n_t$ & number of time steps in PDE solver \\
\midrule
CFL & Courant-Friedrichs-Lewy (number/condition) \\
FD & finite differences \\
FFT & Fast Fourier Transform \\
IP & scattered data interpolation \\
LDDMM & Large Deformation Diffeomorphic Metric Mapping \\
MPI & Message Passing Interface \\
PCG & Preconditioned Conjugate Gradient (method) \\
\bottomrule
\end{tabular}
\end{table}

\subsection{Formulation}\label{s:formulation}

We summarize our notation in \tabref{t:notation-and-symbols}. \claire{} uses an optimal control formulation. We parameterize the deformation map $\vect{y}(\vect{x})$ through a smooth, stationary velocity field $\vect{v}(\vect{x})$. The optimization problem is: Given two images $m_0(\vect{x})$ (template image; image to be deformed) and $m_1(\vect{x})$ (reference image), we seek a \emph{stationary} velocity field $\vect{v}(\vect{x})$ by solving
\begin{subequations} \label{e:problem}
\begin{equation}
\label{e:varopt:objective}
\minopt_{\vect{v}, m}\;\;
\half{1}\!\int_{\Omega}\!(m(\vect{x},1) - m_1(\vect{x}))^2\!\dx
+\frac{\beta_v}{2} \operatorname{reg}_v(\vect{v}) + \frac{\beta_w}{2} \operatorname{reg}_w(w)
\end{equation}

\noindent subject to
\begin{align}
\p_t m(\vect{x},t) + \vect{v}(\vect{x}) \cdot \igrad m(\vect{x},t)
&= 0 &&\text{in}\;\Omega \times(0,1],\label{e:varopt:state}\\
m(\vect{x},t)
&= m_0(\vect{x}) &&\text{in}\;\Omega \times\{0\},\label{e:varopt:constraint}\\
\idiv \vect{v}
&= w &&\text{in}\;\Omega,\label{e:varopt:divconstraint}
\end{align}
\end{subequations}

\noindent on a rectangular domain $\Omega\subset\ns{R}^3$ with periodic boundary conditions on $\p\Omega$. The first term in~\eqref{e:varopt:objective} is a squared $L^2$ image similarity metric, which measures the \textit{distance} between the deformed template image $m(\vect{x},t=1)$ and the reference image $m_1(\vect{x})$. The objective functional~\eqref{e:varopt:objective} additionally consists of two regularization models that act on the controls $\vect{v}$ and $w$ with regularization parameters $\beta_v>0$ and $\beta_w>0$, respectively. The regularization operators are introduced to prescribe sufficient regularity requirements on $\vect{v}$ and its divergence $\idiv \vect{v}$. Smoothness of the velocity guarantees that the computed map is diffeomorphic~\cite{Beg:2005a,Trouve:1998a,Younes:2010a}. We refer to~\cite{Mang:2016H1} for details about our regularization scheme. The default configuration of \claire{} is an $H^1$-Sobolev-seminorm for $\vect{v}$ and $H^1$-Sobolev-norm for $w$~\cite{Mang:2016H1,Mang:2018CLAIRE}. The transport equation~\eqref{e:varopt:constraint} represents the geometrical deformation of $m_0(\vect{x})$ by advecting the intensities forward in time.

To solve~\eqref{e:problem}, we apply the method of Lagrange multipliers to obtain the Lagrangian functional
\begin{equation}
\begin{aligned}
\fun{L}(\vect{\phi}) \defeq &
\half{1}\int_{\Omega}\!(m(\vect{x},1) - m_1(\vect{x}))^2\dx
+\frac{\beta_v}{2} \operatorname{reg}_v(\vect{v}) + \frac{\beta_w}{2} \operatorname{reg}_w(\idiv \vect{v})\\
&
+ \int_0^1\!\!\int_\Omega\lambda(\vect{x},t) (\p_t m + \vect{v} \cdot \igrad m )\dx\dt \\
& + \int_\Omega\lambda(\vect{x},0) (m(\vect{x},0) - m_0(\vect{x})) \dx
+ \int_\Omega p(\vect{x})(\idiv \vect{v} - w) \dx
\end{aligned} \label{eq:Lag}
\end{equation}

\noindent with state, adjoint, and control variables $(m,\lambda,p,\vect{v},w) \defeq \vect{\phi}$, respectively.

\subsection{Discretization and Numerical Algorithms}\label{s:disc_num_algo}

\paragraph{Optimality Conditions \& Reduced Space Approach} To derive the first order optimality conditions, we take the variations of $\fun{L}$ with respect to the state variable $m$, the adjoint variables $\lambda$ and $p$, and the control variable $\vect{v}$. This results in a set of coupled, hyperbolic-elliptic PDEs in \emph{4D (space-time)}. \claire{} uses a reduced-space approach, in which one iterates only on the reduced-space of $\vect{v}$. We require $\vect{g}(\vect{v}^\star) = \vect{0}$ for an admissible solution $\vect{v}^\star$, where
\begin{equation}
\label{e:reducedgrad}
\vect{g}(\vect{v}) \defeq \beta_v \dop{A}\vect{v}(\vect{x}) + \dop{K} \int_0^1 \!\!\!\lambda(\vect{x},t)\igrad m(\vect{x},t)\dt
\end{equation}

\noindent is the so-called reduced gradient. The operator $\dop{A}$ corresponds to the first variation of the regularization model for $\vect{v}$ (i.e., $\operatorname{reg}_v$ in \eqref{e:varopt:objective}) and the operator $\dop{K}$ projects $\vect{v}$ onto the space of near-incompressible velocity fields (see~\cite{Mang:2016H1} for details). To evaluate~\eqref{e:reducedgrad}, we first solve the forward problem~\eqref{e:varopt:constraint} and then the \emph{adjoint problem} given by
\begin{equation}\label{e:adj-transport}
-\p_t \lambda(\vect{x},t) - \idiv \lambda(\vect{x},t)\vect{v}(\vect{x}) = 0 \quad \text{in } \Omega\times[0,1)
\end{equation}

\noindent with final condition $\lambda(\vect{x},t) =  m_1(\vect{x}) - m(\vect{x},t)$ in $\Omega\times\{1\}$ and periodic boundary conditions on $\p\Omega$.

\paragraph{Discretization} We discretize the forward and adjoint PDEs in the space-time interval $\Omega\times[0,1]$, $\Omega \defeq [0,2\pi)^3\subset\ns{R}^3$, with periodic boundary conditions on $\p\Omega$, on a regular grid with $N = N_1N_2N_3$ grid points $\vect{x}_{ijk}\in\ns{R}^3$ in space and $n_t+1$ grid points in time. We use a semi-Lagrangian time-stepping method to solve the transport equations that materialize in the optimality system~\cite{Mang:2017SL,Mang:2016SC}.  Key computational subcomponents of this scheme are $2^{\mathsf{nd}}$-order Runge--Kutta time integrators and spatial interpolation kernels~\cite{Mang:2017SL,Mang:2016SC,Brunn:2020b,Mang:2018CLAIRE}.

To solve the transport equation~\eqref{e:adj-transport} and to evaluate the reduced-gradient $ \vect{g} $~\eqref{e:reducedgrad}, we need to apply gradient and divergence operators. We use an $8^{\mathsf{th}}$ finite difference (\acr{FD}) scheme for these first-order differential operators~\cite{Brunn:2020a,Brunn:2020b}. The reduced gradient~\eqref{e:reducedgrad} also involves the vector-Laplacian $\dop{A}$ and the Leray-like operator $\mathcal{K}$ (see \cite{Mang:2016H1}). In spectral methods, inversion and application of higher-order differential operators come at the cost of two FFTs and one Hadamard product in Fourier space.

\paragraph{Gauss--Newton--Krylov Solver}

\claire{} uses a Gauss--Newton--Krylov method globalized with an Armijo line search to solve the non-linear problem $\vect{g}(\vect{v}) = \vect{0}$~\cite{Mang:2015NK,Mang:2018CLAIRE}. The iterative scheme is given by
\begin{equation}
\label{e:iter}
\vect{v}_{k+1} = \vect{v}_k + \alpha_k \vect{\tilde{v}}_k,
\quad \mat{H}\vect{\tilde{v}}_k = -\vect{g}_k,
\quad k = 0,1,2,\ldots,
\end{equation}

\noindent where $\mat{H}\in\ns{R}^{3N,3N}$ is the discretized reduced-space Hessian operator, $\vect{\tilde{v}}_k\in\ns{R}^{3N}$ the search direction, $\vect{g}_k\in\ns{R}^{3N}$ a discrete version of the gradient in~\eqref{e:reducedgrad}, $\alpha_k>0$ a line search parameter, and $k\in\ns{N}$ the Gauss--Newton iteration index. We have to solve the linear system in~\eqref{e:iter} at each Gauss--Newton step. We do not form or assemble $\mat{H}$; we use a matrix-free preconditioned conjugate gradient (\acr{PCG}) method to solve $\mat{H}\vect{\tilde{v}}_k = -\vect{g}_k$ for $\vect{\tilde{v}}_k$. This only requires an expression for applying $\mat{H}$ to a vector that we term Hessian matvec. In continuous form, the Gauss--Newton approximation of this matvec is given by
\begin{equation}
\label{e:matvec}
\dop{H}\vect{\tilde{v}} =
\beta_v\dop{A}\vect{\tilde{v}}(\vect{x})
+ \dop{K} \int_0^1\tilde{\lambda}(\vect{x},t)\igrad m(\vect{x},t) \dt.
\end{equation}

Similarly to the evaluation of the reduced gradient in \eqref{e:reducedgrad}, the application of the Hessian to a vector in \eqref{e:matvec} requires the solution of two PDEs to find the space-time field $\vect{\tilde{v}}$ (see \cite{Mang:2015NK,Mang:2016H1,Mang:2018CLAIRE} for details). Consequently, solving the linear system with $\mat{H}$ in \eqref{e:iter} is the most expensive part of \claire{}. Preconditioning of the reduced-space Hessian system can be used to alleviate these computational costs. In~\cite{Brunn:2020b} we have introduced a zero velocity approximation for $\mat{H}$ as a preconditioner. This preconditioner can be applied at full resolution and through a two-level coarse grid approximation (see \cite{Brunn:2020b} for details). The latter variant represents the default considered in the present work.

\section{Computational Kernels and Parallel Algorithms}
\label{sec:kernels}

At each Gauss--Newton step, we have to solve the forward and the adjoint equations for the reduced gradient and the Hessian matvecs. The main computational cost in \claire{} constitute FFTs for (inverse) differential operators, scattered data interpolation (\acr{IP}) for the semi-Lagrangian solver, and FD for computing first order derivatives (see~\cite{Mang:2017SL,Mang:2016SC,Brunn:2020a,Brunn:2020b} for a detailed description of these computational components). The distributed memory CPU implementation of \claire{} uses \texttt{AccFFT}~\cite{accfft-home-page,accfft_github} for spectral operations~\cite{Mang:2016SC,Gholami:2017SC}. In the single GPU setup, we use the highly optimized 3D FFT operations provided by NVIDIA's \texttt{cuFFT} library. In the multi-node multi-GPU setup, we use a 2D slab decomposition to leverage 2D \texttt{cuFFT} functions. We decompose the spatial domain in $x_1$ direction, which is the outer-most dimension, and the spectral domain in the $x_2$ direction. Let $p$ be the number of MPI tasks. Then, each MPI task gets $(N_1/p) \times N_2 \times N_3$ grid points, where $N_1, N_2, N_3$ are the image dimensions. We have discussed the implementation details and shown scalability of the FFT kernel in~\cite{Brunn:2020b}.

The parallel implementation of our IP kernel on CPUs was introduced in~\cite{Mang:2016SC} and improved in~\cite{Gholami:2017SC}. In~\cite{Brunn:2020a}, we explored linear, cubic Lagrange, and cubic B-spline interpolation schemes for the interpolation kernel on a single GPU setup. In~\cite{Brunn:2020b}, we ported these kernels to the multi-node multi-GPU setup and made several optimizations. In the present study, we use linear interpolation to evaluate the image intensities at the off-grid points (also called characteristic points) in our semi-Lagrangian scheme. Depending on the image data layout and the velocity field, the IP kernel requires scattered peer-to-peer communication of off-grid points between the owner and the worker processors.

The CPU version of \claire{} uses FFTs for spatial derivatives~\cite{Mang:2018CLAIRE, Mang:2016SC, Gholami:2017SC}. In~\cite{Brunn:2020a}, we introduced the $8^{\mathsf{th}}$ order FD kernel to evaluate first order derivatives, i.e., spatial gradients and divergence operators on a single GPU. In~\cite{Brunn:2020b}, we ported the FD kernel to the multi-GPU setup. We use the FD kernel for computing first order derivatives throughout the registrations performed in this paper.

\claire{} uses CUDA-aware MPI in the multi-node multi-GPU setup, thereby avoiding unnecessary CPU-GPU communication and automatically utilizing the high-speed on-node \textit{NVLink} interconnect bus between GPUs if it is available.

\subsection{Compute Hardware and Libraries}

All runs reported in this study were executed on TACC's Longhorn system in single precision. Longhorn hosts 96 NVIDIA Tesla V100 nodes. Each node is equipped with four GPUs and 16 GB GPU RAM each (i.e., 64 GB per node) and two IBM Power 9 processors with 20 cores (40 cores per node) at 2.3 GHz with 256 GB memory. Our implementation uses \texttt{PETSc}~\cite{petsc-web,petsc-efficient} for linear algebra, the \texttt{PETSc} \texttt{TAO} package for nonlinear optimization, CUDA~\cite{cuda-web}, thrust~\cite{Thrust}, \texttt{cuFFT} for FFTs~\cite{Nvidia2007b}, \texttt{niftilib}~\cite{niftilib-web} for serial I/O for small images and \texttt{PnetCDF}~\cite{pnetcdflib} for parallel I/O for large scale images, IBM Spectrum MPI~\cite{ibmspectrum-web}, and the IBM XL compiler~\cite{ibmxl-web}.

\subsection{Code availability}
\claire{}~\cite{claire-web, Brunn2021} is available publicly for download on github at
\begin{center}
\url{https://github.com/andreasmang/claire}
\end{center}

under the GNU General Public License v3.0.

\subsection{Key Solver Parameters}
Here, we summarize the key parameters of \claire{} and discuss their effect on the solver and previous strategies to choose suitable values. In \secref{s:param_cont}, we present our algorithm to choose these parameters in a combined continuation approach.\\[1mm]

\noindent \textbf{$\beta_v$ --- regularization parameter for the velocity field $\vect{v}$.} Large values for $\beta_v$ result in very smooth velocities and, thus, maps that are typically associated with a large final image mismatch. Smaller values of $\beta_v$ allow complex deformations but lead to a solution that might be close to being non-diffeomorphic due to discretization issues. From a user application point of view, we are interested in computing velocity fields, for which the Jacobian determinant, i.e., the determinant of the deformation gradient $\mat{F} \coloneqq \nabla \vect{y} $, is strictly positive for every image voxel. This guarantees a locally diffeomorphic transformation (subject to numerical accuracy). In~\cite{Mang:2015NK,haberGCVBasedMethod2000}, we determined the regularization parameter $\beta_v$ based on a binary search algorithm. The search is constrained by the bounds on $J = \det\mat{F}$. That is, we choose $\beta_v$ such that $J$ is bounded from below by $J_{\text{min}}$ and bounded from above by $1/J_{\text{min}}$, where $J_{\text{min}}\in(0,1)$ is a user-defined parameter. The binary search is expensive because we solve the inverse problem repeatedly: For each trial $\beta_v$, we iterate until the convergence criteria for the Gauss--Newton--Krylov solver is met then use the previous velocity field as an initial guess for the next trial $\beta_v$.\\[1mm]

\noindent \textbf{$\beta_w$ --- regularization parameter for the divergence of the velocity field $w=\nabla \cdot \vect{v}$.} The choice of $\beta_w$, along with $\beta_v$, is equally critical. Small values can result in extreme values of $J$ and make the deformations locally non-diffeomorphic. As discussed above, in our previous work~\cite{Mang:2015NK}, we do parameter continuation in $\beta_v$ and keep $\beta_w$ fixed. This is sub-optimal for two reasons: \emph{(i)} Both $\beta_v$ and $\beta_w$ depend on the resolution, so keeping $\beta_w$ fixed for all resolutions can result in deformations with undesirable properties, and \emph{(ii)} doing continuation in $\beta_v$ alone does not ensure we get close enough to the set Jacobian bounds. Therefore, adding continuation in $\beta_w$, which also affects the Jacobian, is necessary.\\[1mm]

\noindent \textbf{$J_{\text{min}}$ --- lower bound for the determinant $J$ of the deformation gradient.} The choice of this parameter is typically driven by dataset requirements, i.e., one has to decide how much volume change is acceptable. \claire{} uses a default value of 0.25~\cite{Mang:2018CLAIRE}. Tighter bound on the Jacobian, i.e., $J_{\text{min}}$ close to unity, will result in large $\beta_v$ and $\beta_w$ values leading to simple deformations and sub-par registration quality. Relaxing the Jacobian bound in combination with our continuation schemes for $\beta_v$ and $\beta_w$ can result in very small regularization parameters and extremely complex deformations.\\[1mm]

\noindent \textbf{$n_t$ --- number of time steps in the semi-Lagrangian scheme}. The semi-Lagrangian scheme is unconditionally stable and outperforms RK2 time integration schemes in terms of runtime for a given accuracy tolerance~\cite{Mang:2017SL}. The choice of $n_t$ is based on the adjoint error, which is the error measured after solving~\eqref{e:varopt:constraint} forward and then backward in time. In~\cite{Mang:2017SL}, we conducted detailed experiments for 2D image registration and found, that even for problems of clinical resolution $n_x=256^2$, $n_t=3$ (CFL=$10$) did not cause issues in solver convergence. Increasing $n_t$ beyond a certain value will introduce additional discretization errors from the interpolation scheme. \\[1mm]

\noindent \textbf{Resolution of $\vect{v}$}. We use the same spatial discretization for $\vect{v}$ as given for the input images. There exist image registration algorithms that approximate the registration deformation in a low-dimensional bandlimited space without sacrificing accuracy, resulting in dramatic savings in computational cost~\cite{Zhang:2019a}. We have not explored this within the framework of \claire{}. Note that~\cite{Zhang:2019a} uses higher order regularization operators, which leads to smoother velocities compared to the ones \claire{} produces, therefore enabling a representation on a coarser mesh. Moreover, \claire{} uses a stationary velocity field, i.e., $\vect{v}$ is constant in time. In our previous work~\cite{Mang:2015NK}, we have demonstrated that stationary and time-varying velocity fields yield similar registration accuracy for registration between two real medical images of different subjects. More precisely, we did not observe any practically significant quantitative differences in registration accuracy for a varying number of coefficient fields in the case of time-varying velocity fields. Using a stationary velocity field is significantly cheaper and has a smaller memory overhead from a computational cost perspective.

\subsection{Parameter Identification}\label{s:param_cont} Our algorithm to choose solver parameters proceeds as follows.\\[1mm]

\noindent \textbf{Resolution-dependent choice of the interpolation order and $n_t$.} As our GPU implementation is only available in single precision (unlike the CPU implementation~\cite{Mang:2018CLAIRE}, which is available both in single and double precision), we use cubic interpolation (B-splines/Lagrange polynomials) with $n_t=4$ ($n_t=8$ for linear interpolation) for resolutions up to $\vect{n}_x=(256,256,256)$. For higher resolutions, we use linear interpolation to save computational cost and increase $n_t$ proportionately to $n_x$ to keep the CFL number fixed.\\[1mm]

\noindent \textbf{Parameter search scheme for $\beta_v$ and $\beta_w$.} We perform a two-stage search scheme: \emph{(i)} In the first part of the parameter search, we fix $\beta_w = \beta_{w,\text{max}}$ ($\beta_{w,\text{init}} = 1\text{e-}05$) and search for $\beta_v$. The registration problem is first solved for a large value of $\beta_v = \beta_{v,\text{max}}$ so that we under-fit the data. In our experiments, we set $\beta_{v,\text{max}}=1$. Subsequently, $\beta_v$ is reduced by one order of magnitude in every continuation step and the registration problem is solved again with the new $\beta_v$. We repeat the reduction of $\beta_v$ until we breach the Jacobian bounds $[J_{\text{min}}, 1/J_{\text{min}}]$. When this happens, we do a binary search for $\beta_v$ between the last two values and terminate the binary search when the relative change in $\beta_v$ is less than 10\% of the previous valid $\beta_v$. In addition, we put a lower bound $\beta_{v,\text{min}}=1\text{e-}05$ on $\beta_v$. This lower bound is set purely to minimize computational cost. We denote the final value of $\beta_v$ as $\beta_v^\star$.

\emph{(ii)} In the second part of the search, we do a simple reduction search for $\beta_w$ by fixing $\beta_v=\beta_v^\star$. Starting with a given value $\beta_{w,\text{max}}$, we reduce $\beta_w$ by one order of magnitude and repeat solving the registration problem with $\beta_v^\star$ and the respective value for $\beta_w$ until we reach $J_{\text{min}}$. We put a lower bound $\beta_{w,\text{min}}=1\text{e-}07$ on $\beta_w$ in order to minimize computational cost. We take the last valid value of $\beta_w$, for which the Jacobian determinant was within bounds and denote it as $\beta_w^\star$. We fixed the value of $\beta_{w,\text{max}}=1\text{e-}05$ for all experiments and resolutions. We determined this value empirically by running image registration on a couple of image pairs at resolution $640\times880\times880$ and $160\times220\times220$ (see~\secref{exp_1b} for the images) for different values of $\beta_{w,\text{max}}$. We report these runs in~\tabref{tab:brainMRI250um_nirep_velocity_multilevel_Dice_stats_iporder1_betaw_variation} (see~\appref{a:determine_betaw}).

We evaluate the parameter search scheme for real world brain images and report the performance in~\secref{s:param_cont_results}. Furthermore, we use it as the default parameter search scheme for all the experiments presented in this paper.\\[1mm]

\noindent \textbf{Parameter continuation scheme for $\beta_v$ and $\beta_w$.} If we want to use target $\beta_v^\star$ and $\beta_w^\star$ values for a new registration problem, we can perform a parameter \textit{continuation} which is exactly like the parameter \textit{search} except that we neither perform the binary search for $\beta_v$ nor check for the bounds on $J$. In the first stage of the \textit{continuation}, we solve the registration problem for successively smaller values of $\beta_v$ starting from $\beta_v = 1$ and reducing it by one order of magnitude until we reach $\beta_v=1\text{e}k $ where $k = \lceil\log_{10}(\beta^\star_v)\rceil $. Then we do an additional registration solve at $\beta_v =\beta_v^\star$. We fix $\beta_w=\beta_{w,\text{max}}$ in the first stage. In the second stage, we fix $\beta_v = \beta_v^\star$ and reduce $\beta_w$ from $\beta_{w,\text{max}}$ to $\beta_w^\star$ in steps of one order of magnitude.

Whereas the expensive parameter search allows us to identify an optimal set of regularization parameters for unseen data, we use the parameter continuation scheme to speed up convergence. The combination of both is particularly efficient, for example in cohort studies, where we identify optimal regularization parameters for \textit{one} image pair in the cohort and use the obtained parameters for all the other images.

\section{Materials}\label{s:datasets}
We use publicly available image datasets for carrying out the image registration experiments in this paper (see~\secref{s:results}). We summarize these datasets in~\tabref{tab:datasets}. We discuss these datasets in detail.

\paragraph{MUSE} This dataset consists of five real brain $T_1$-weighted MRIs of different individuals. These images were segmented into 149 functional brain regions (see~\figref{fig:muse_claire}) in a semi-automated manner including manual corrections by expert radiologists~\cite{muse_templates}. These images are part of a bigger set of template images that were used for the development of the MUSE~\cite{Doshi:2016a} segmentation algorithm. The original image size is $ 256\times256\times256$ at a spatial resolution of 1~mm. This dataset is available for download through the neuromorphometrics website~\cite{muse_templates}.

\paragraph{NIREP} \cite{Christensen:2006a} is a standardized repository for assessing registration accuracy that contains 16 $T_1$-weighted MR neuroimaging datasets (\texttt{na01}--\texttt{na16}) of different individuals at an isotropic resolution of 1~mm. The original image size is $256\times300\times256$ voxels. We resample these images to an isotropic image size of $ 256\times256\times256 $. We use the images \texttt{na01}-\texttt{na10} for our experiments. This dataset is available for download through the \texttt{GitHub} link  https://github.com/andreasmang/nirep.

\paragraph{SYN} We create four sets of synthetic template and reference images to assess image registration accuracy as a function of resolution. We create a set of synthetic reference images $m_1$ by solving~\eqref{e:varopt:constraint} using a given synthetic template image $m_0$ and a synthetic velocity field $\vect{v}$. To construct the template image $m_0$, we use a linear combination of high-frequency spherical harmonics. To be precise, we define the template image $m_0(\vect{x})$ as
\begin{subequations}
\begin{equation}
\label{e:geeee}
m_0(\vect{x}) = \sum_{i=1}^{10} g_i(\vect{x})
\quad \text{with}\qquad
g_i(\vect{x}) =
\begin{cases}
1, & \text{if } \|\vect{x} - \hat{\vect{x}}_i\|_2 \leq | Y_l^m( \theta + \hat{\theta}_i, \phi + \hat{\phi}_i)|, \\
0, & \text{otherwise},
\end{cases}
\end{equation}

\noindent and image coordinates $\vect{x} \coloneqq (x,y,z) \in (-\pi,\pi]^3$. In \eqref{e:geeee}, $Y_l^m$ represents spherical harmonics of the form
\begin{equation}
Y_l^m(\theta, \phi) = \sqrt{\frac{2l+1}{4\pi} \frac{(l-m)!}{(l+m)!}} e^{im\theta}P_l^m (\cos(\phi))
\end{equation}
\end{subequations}

\noindent with parameters $m$, $l$, angular directions $\theta \in [0,\pi]$ and $\phi \in [0,2\pi]$, and associated Legendre functions $P^m_l$. We choose $m=6$, $l=8$ for our setup. $\hat{\theta}_i$ and $\hat{\phi}_i$ are random perturbations in integer multiples of $\pi/2$ and $\hat{\vect{x}}_i \in [-0.4\pi, 0.4\pi]^3$ is a random offset from the origin. The reference image $m_1(\vect{x})$ is generated by solving~\eqref{e:varopt:constraint} with initial condition $m_0(\vect{x})$ and velocity field $\vect{v}(\vect{x}) \defeq (v_x(\vect{x}), v_y(\vect{x}), v_z(\vect{x}))$, $\vect{x} = (x,y,z)$, defined as
\begin{equation}
\label{e:syn_vel}
v_x = \sum_{k=1}^{K} \frac{1}{k^{0.5}}\cos(ky)\cos(kx), \quad
v_y = \sum_{k=1}^{K} \frac{1}{k^{0.5}}\sin(kz)\sin(ky), \quad
v_z = \sum_{k=1}^{K} \frac{1}{k^{0.5}}\cos(kx)\cos(kz),
\end{equation}

\noindent where $K = \{4,8,12,16\}$. We set the template and the reference base image size to $\vect{n}_x = \vect{n} = (1024,1024,1024)$. It is important to note that $m_0$ and $m_1$ possess only the discrete intensities $i \in \{1,2,\ldots,10\}$. This allow us to naturally define ten labels $l_0^i$ and $ l_1^i $ corresponding to $m_0$ and $m_1$, respectively, for all image voxels with intensity $i$ for each $i \in \{1,2,\ldots,10\}$. We show a 2D slice of the template $m_0$ and reference $m_1$ images for the case $K=4$ in~\figref{fig:synthetic-spherical-harmonics-K-8_sl2}. The scripts for generating the template image $m_0$ and the synthetic velocity field $\vect{v}$ can be found at \url{https://github.com/naveenaero/scala-claire}. The reference image $m_1$ can be generated using \claire{}~\cite{claire-web,Brunn2021}.

\paragraph{MRI250} \cite{Lusebrink:2017a} is an \emph{in-vivo} 250~$\mu$m human brain MRI image which consists of a $T_1$-weighted anatomical data acquired at an isotropic spatial resolution of 250~$\mu m$. The original image size is $640\times880\times880$ voxels. This image can be downloaded from~\cite{mri250}. We skull strip the dataset by downsampling it to $128\times128\times128$ using linear interpolation and then manually create the brain mask in ITK-SNAP~\cite{itksnap}. We upsample this brain mask back to the original resolution and then apply it to the original image. We use the tool \texttt{fast}~\cite{Zhang:2001a} from the FSL toolkit~\cite{Woolrich:2009a,Smith:2004a,jenkinson2012fsl} to segment the $T_1$-weighted MRI into gray matter (GM), white matter (WM) and cerebrospinal fluid (CSF) to be able to evaluate the registration performance using Dice score (see~\eqref{e:dice}) between the image labels before and after registration.

\paragraph{CLARITY} We use the dataset from~\cite{Chung:2013a, Vogelstein:2018b,Kutten:2016a,Kutten:2017b} which consists of 12 mouse brains images acquired using CLARITY-Optimized Light-sheet Microscopy (\acr{COLM}). This dataset is available for download from~\cite{clarity-web}. These images have low contrast and are noisy. The in-plane resolution is $0.585\mu\text{m} \times 0.585\mu\text{m}$ and the cross-plane resolution is 5 to 8~$\mu$m. The images are stored at eight different resolution levels with level zero being the full resolution and level seven being the lowest resolution. We use the images at resolution levels three and six in our experiments. These levels correspond to an in-plane resolution of $4.68 \mu\text{m} \times 4.68\mu\text{m}$ and $37.44\mu\text{m}\times 37.44\mu\text{m}$, respectively, which translates to images of size $n=(2816,3016)$ and $n/8=(328,412)$ voxels. The cross-plane resolution is constant at all levels and corresponds to 1162 voxels. We select \texttt{Control182}, \texttt{Fear197} and \texttt{Cocaine178} as the test images in our experiments.

\begin{table}
\centering
\caption{We list the image datasets we use in our scalable registration experiments (see~\secref{s:results}). All the datasets are accessible publicly and further discussed in~\secref{s:datasets}. We list the dataset name tag (which we use to refer to them throughout the rest of the paper), the imaging modality, the number of images, the spatial resolution and the image resolution in voxels. For dataset with an isotropic spatial resolution, we only provide a single value. For datasets with anisotropic spatial resolution, we list the resolution in all three dimensions. For the \textbf{SYN} dataset, spatial resolution does not carry a physical meaning, so we only list the image resolution.}\label{tab:datasets}
\small
\begin{tabular}{ccccc}
\toprule
\bf dataset & \bf image modality & \bf number of images & \bf spatial resolution & \bf image resolution\\
\midrule
\textbf{MUSE} & $ T_1 $-weighted MRI & 5 & 1~mm & (256,256,256) \\
\textbf{NIREP} & $ T_1$-weighted MRI & 16 & 1~mm & (256,300,256) \\
\textbf{SYN} & synthetic & 4 & -- & (1024,1024,1024) \\
\textbf{MRI250} & $ T_1 $-weighted MRI & 1 & 250$~\mu$m & (640,880,880) \\
\textbf{CLARITY} & \begin{tabular}{@{}c@{}}CLARITY-optimized \\ light sheet microscopy\end{tabular} & 3 & (4.68,4.68,5)~$ \mu $m & (2816,3016,1162) \\
\bottomrule
\end{tabular}
\end{table}

\section{Results and Discussion}\label{s:results}

We test the image registration on real-world (see~\secref{exp_1b} and \secref{exp_2}) and synthetic registration problems (see~\secref{exp_1a}). The measures to analyze the registration performance are summarized in~\secref{s:perf_measure}. We evaluate the parameter search scheme (see~\secref{s:param_cont}) on a set of real brain images and present the results in~\secref{s:param_cont_results}. Furthermore, we explore the following questions in the context of scalable image registration:\\[1mm]

\noindent\textbf{Question Q1:} Do we need large scale high resolution image registration? Does the registration quality degrade when the registration is performed at a downsampled resolution when compared to performing registration at the original high resolution?\\[1mm]

\noindent\textbf{Question Q2:} How does registration perform and scale for real, noisy and high resolution medical images of human and mouse brains?\\[1mm]

\subsection{Measures of Performance}\label{s:perf_measure}
In our experiments, we evaluate both runtime performance (in terms of solver wall clock time) and the registration quality in terms of accuracy. For the latter, we use the following metrics:

\subsubsection*{Dice Score Coefficient $D$.} Let $l_0$ and $l_1$ be the binary label maps associated with the images $m_0$ and $m_1$, respectively. Then the Dice score $D$ between the two label maps is given by
\begin{equation}\label{e:dice}
D(l_0, l_1) = \frac{2 |l_0 \cap l_1|}{|l_0| + |l_1|},
\end{equation}

\noindent where $|\cdot|$ denotes the cardinality of a set, and $\cap$ denotes the intersection of the two sets, respectively. We define $D(l_0,l_1)$ to be the Dice score pre-registration and $D(l(t=1),l_1)$ post-registration, where $l(t=1)$ is the label map that corresponds to the deformed template image $m(t=1)$. Furthermore, for a set of discrete labels $l^i$, $i=\{1,2,\ldots,M\}$, where $i$ corresponds to the label index, we define the volume fraction
\begin{equation*}
\alpha^i = \frac{|l^i|}{\sum_{i=1}^{M} |l^i|}.
\end{equation*}

\noindent Using this definition, we compute the following statistics for the Dice coefficient: The Dice coefficient average $D_a$ given by
\begin{equation}\label{e:dice_avg}
D_a = \frac{1}{M}\sum_{i=1}^{M} D(l_0^i, l_1^i) ,
\end{equation}

\noindent the volume weighted average of the Dice coefficient given by
\begin{equation}\label{e:dice_vol_avg}
D_{vw} = \frac{1}{\sum_{i=1}^{M}|l_1^i|} \sum_{i=1}^{M} |l_1^i| D(l_0^i, l_1^i),
\end{equation}

\noindent and the inverse of the volume weighted average Dice coefficient given by
\begin{equation}\label{e:dice_inv_vol_avg}
D_{ivw} = \frac{1}{\sum_{i=1}^{M} 1/|l_1^i|} \sum_{i=1}^{M} \frac{D(l_0^i, l_1^i)}{|l_1^i|}.
\end{equation}

\noindent Note that $D_{vw}$ gives more weight to labels with higher volume fractions while $D_{ivw}$ gives more weight to labels with smaller volume fractions.

\subsubsection*{Relative Residual $r$.} This metric corresponds to the ratio of the image mismatch before and after the registration. It is given by
\begin{equation}
r = \frac{||m(t=1) - m_1||^2_2}{||m_0 - m_1||^2_2}.
\end{equation}

\subsubsection*{Characteristic Parameters.} For each image registration, we also report the regularization parameters and the obtained minimum and maximum values of the determinant of the deformation gradient $J \defeq \det \mathbf{F}$, i.e., the determinant of the Jacobian of the deformation map.

\subsubsection*{Visual Analysis.} We visually support this quantitative analysis with snapshots of the registration results. The registration accuracy can be visually judged from the residual image, which corresponds to the absolute value of the pointwise difference between $m(t=1)$ and $m_1$. The regularity of the deformations can be assessed from the pointwise maps of the determinant of the deformation gradient.

\subsection{Experiment 1: Evaluation of the Parameter Search Scheme}\label{s:param_cont_results}

We evaluate the parameter search scheme on a set of real brain images and compare the registration performance with a state-of-the-art \texttt{SyN} deformable registration tool in the \texttt{ANTs} toolkit.

\paragraph{Dataset} We use the \textbf{MUSE} dataset (see~\secref{s:datasets}) for this experiment. After registration of the original $T_1$-weighted images from this dataset, we use the image labels to evaluate the registration performance in terms of the volume weighted average Dice score $D_{vw}$.

\paragraph{Procedure} Out of the five $T_1$ images, we select \texttt{Template27} as the reference image $m_1$ and register the other four images to $m_1$. For the registration, we use the parameter search scheme (see~\secref{s:param_cont}) to identify best regularization. We use linear interpolation and $n_t=8$ time steps in the semi-Lagrangian solver. For the Jacobian bound, we select $J_{min}=0.1$. In the parameter search, for each trial $\beta_v$ and $\beta_w$, we drive the relative gradient norm $\|g\|_{2,rel}=\|g\|_2/\|g_0\|_2$ to $1\text{e-}02$. Once we have found adequate $\beta_v$ and $\beta_w$ for each image pair, we rerun the image registrations using only parameter continuation. For a baseline performance comparison, we also perform registration on the same image pairs using the \texttt{SyN} tool in  \texttt{ANTs}~\cite{Avants:2008a}. For \texttt{ANTs}, we use the ``MeanSquares'' (i.e., squared $L_2$-) distance measure. We run \claire{} on a single NVIDIA V100 GPU with 16GB of memory on TACC's Longhorn supercomputer. We run \texttt{ANTs} on a single node of the TACC Frontera supercomputer (system specs: Intel Xeon Platinum 8280 (``Cascade Lake'') processor with 56 cores on 2 sockets (base clock rate: 2.7GHz)). We use all 56 cores. We report the parameters used for \texttt{ANTs} in~\secref{s:ants_affine_params}.

\begin{table}
\centering
\caption{\textbf{Experiment 1: Performance of the parameter search scheme implemented in \claire{}}. We report results for the registration of four template images to the reference image \texttt{Template27}. We consider the squared $L_2$-distance measure as image similarity metric. We restrict the Jacobian determinant $J\in[0.1,10]$ for these registrations. We report the following quantities of interest: (i) optimal regularization parameters $\beta_v^\star$ and $\beta_w^\star$, (ii) minimum $J_{\text{min}}$ and maximum $J_{\text{max}}$ Jacobian determinant achieved, (iii) solver wall clock time in seconds, and (iv) label volume weighted Dice average $D_{vw}$ pre and post registration.}\label{tab:muse_claire}
\small
\begin{tabular}{crrrrrrrr}
\toprule
{Template} & $\beta^\star_v$ & $\beta^\star_w$ & $J_{\text{min}}$ & $J_{\text{max}}$ & \multicolumn{2}{c}{$D_{vw}$} & \multicolumn{2}{c}{runtime(s)} \\
{} &   & & & &   pre & post & search & continuation \\

\midrule
4        &  7.75e-05 &  1.00e-04 &  4.53e-01 &  5.36e+00 & 5.53e-01 & 6.99e-01 &   5.90e+02 &     4.04e+01 \\
16       &  7.89e-05 &  1.00e-05 &  2.62e-01 &  4.23e+00 & 5.51e-01 & 6.95e-01 &   4.39e+02 &     5.82e+01 \\
22       &  1.14e-05 &  1.00e-04 &  1.19e-01 &  1.74e+00 & 5.39e-01 & 7.04e-01 &   7.05e+02 &     9.79e+01 \\
31       &  2.83e-05 &  1.00e-04 &  2.40e-01 &  1.86e+00 & 5.27e-01 & 7.00e-01 &   6.19e+02 &     6.07e+01 \\
\bottomrule
\end{tabular}
\end{table}

\begin{table}
\centering
\caption{\textbf{Experiment 1: Performance of \texttt{ANTs}}. We report results for registration of four template images to the reference image \texttt{Template27} using a squared $L_2$-distance metric. We report the following quantities of interest (i) minimum ($J_{\text{min}}$) and maximum ($J_{\text{max}}$) determinant of the deformation gradient obtained, (ii) label volume weighted Dice average $D_{vw}$ \emph{pre} and \emph{post} registration, and (iii) solver wall clock time in seconds.}\label{tab:muse_ants}
\small
\begin{tabular}{crrrrr}
\toprule
{Template} & $J_{\text{min}}$ & $J_{\text{max}}$ & \multicolumn{2}{c}{$D_{vw}$} & runtime(s) \\
{} &    & &   pre & post & \\
\midrule
4   &  1.40e-01 & 3.10e+0   & 5.53e-01 & 6.86e-01 &        1.98e+02 \\
16  &  2.50e-01 & 4.59e+0   & 5.51e-01 & 6.87e-01 &        2.00e+02 \\
22  &  3.11e-01 & 9.73e+0   & 5.39e-01 & 6.62e-01 &        1.99e+02 \\
31  &  2.07e-01 & 4.76e+0   & 5.27e-01 & 6.85e-01 &        2.10e+02 \\
\bottomrule
\end{tabular}
\end{table}

\begin{figure}
\centering
\includegraphics[width=\linewidth]{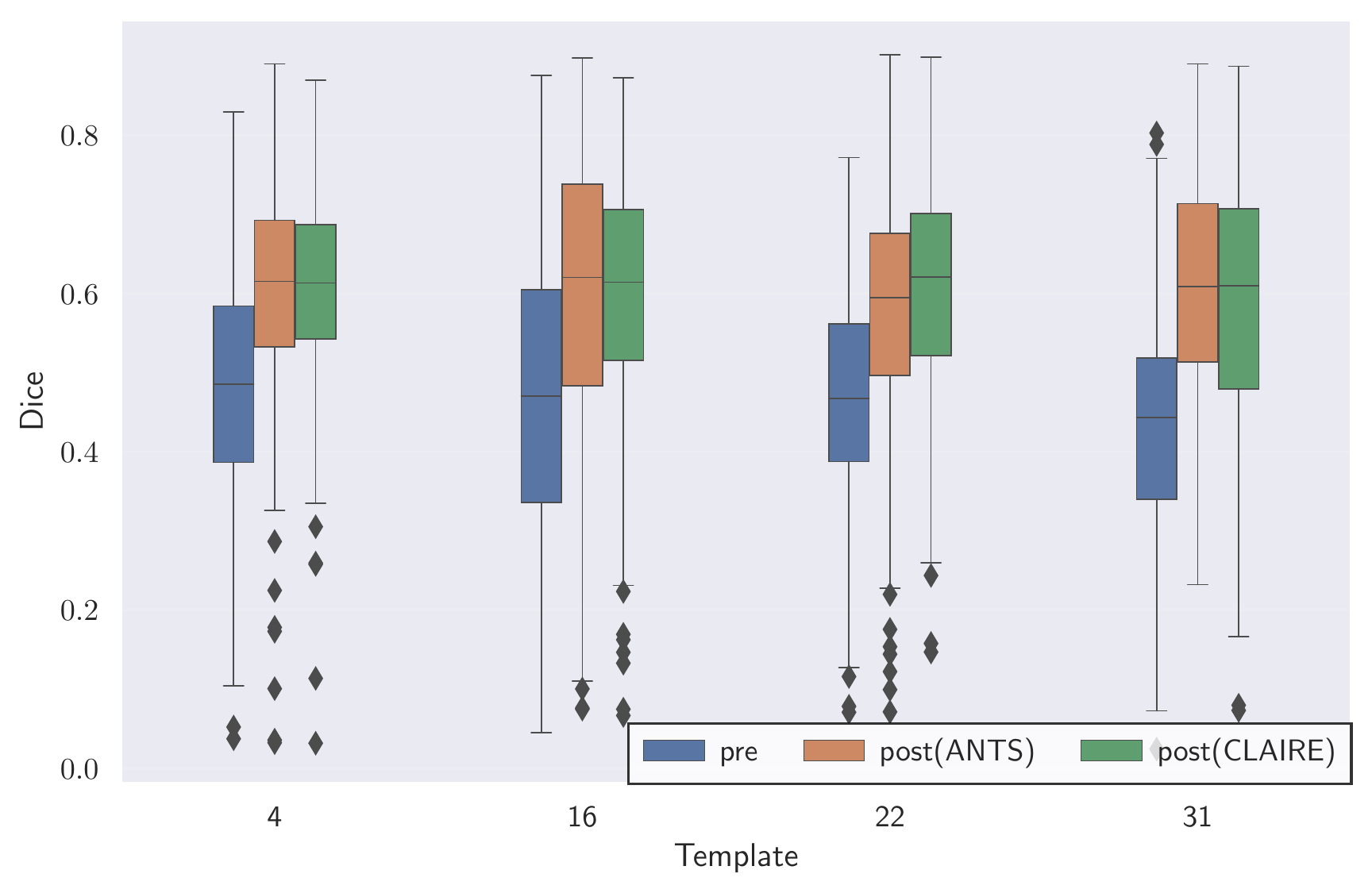}
\caption{\textbf{Experiment 1: Comparison of Dice scores for \claire{} and \texttt{ANTs}}. The box plots show Dice scores of the individual labels for the registration results reported for \claire{} in~\tabref{tab:muse_claire} and \texttt{ANTs} in~\tabref{tab:muse_ants}.
}\label{fig:muse_claire_dice_box_plots}
\end{figure}

\begin{figure}
\centering
\includegraphics[width=\linewidth]{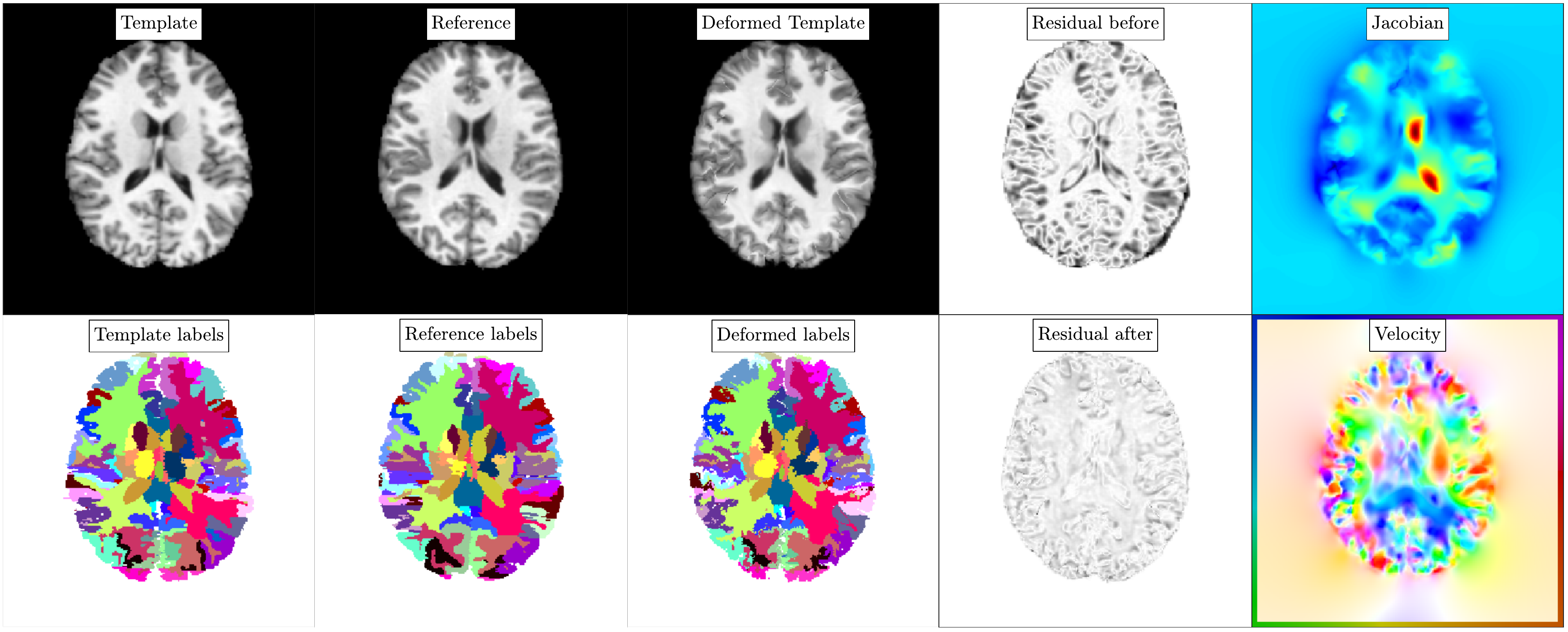}
\caption{\textbf{Experiment 1: Exemplary registration results using the parameter search scheme implemented in \claire{}}. We consider the datasets \texttt{Template16} (template image) and \texttt{Template27} (reference image). We refer to~\tabref{tab:muse_claire} and the text for details about the setup. We show (from left to right) the template, reference, deformed template image (top row) and their corresponding labels (bottom row). We also visualize the residual before and after the registration along with the determinant of the deformation gradient and an orientation map for the velocity field.}\label{fig:muse_claire}
\end{figure}

\paragraph{Results} We report the obtained estimates for $\beta_v$ and $\beta_w$ as well as results for registration quality in~\tabref{tab:muse_claire}. In~\figref{fig:muse_claire}, we provide a representative illustration of the obtained registration results. We report baseline registration performance using \texttt{ANTs} in~\tabref{tab:muse_ants}. We compare the Dice scores obtained for \claire{} and \texttt{ANTs} in~\figref{fig:muse_claire_dice_box_plots}.

\paragraph{Observations} \claire{} allows us to precisely control the properties of the deformation without having to tune any parameters manually. The only free parameters are the Jacobian bounds, which depend on the overall workflow related to the dataset. The volume weighted Dice scores $D_{vw}$ obtained for \claire{} (see~\tabref{tab:muse_claire}) are competitive to those produced by \texttt{ANTs} (see~\tabref{tab:muse_ants}). The average runtime for ANTs for all the registrations reported in~\tabref{tab:muse_ants} is 201\,seconds ($\approx 3$ minutes). For \claire{}, the average wall clock time of \claire{} in the parameter search mode is 9.8\,minutes ($3\times$ slower than \texttt{ANTs}; we search for adequate regularization parameters), while, in the continuation mode, the runtime of \claire{} is 64\,seconds ($3\times$ faster than \texttt{ANTs}; we apply the optimal regularization parameter and do not search for them).

\subsection{Experiment 2A: High Resolution Synthetic Data Registration}\label{exp_1a}

In this experiment, we answer \textbf{Q1}. We attempt this by executing our registration algorithm on synthetic imaging data. The advantages of using such images over real datasets are as follows:
\begin{itemize}
\item They are noise-free, high contrast, and sharp, unlike real-world images.
\item There is a scarcity of high resolution real image data because it is expensive and time-consuming to acquire. We can control the resolution of synthetic data because the images are created using analytically known functions.
\item We can control the number of discrete image intensity levels, i.e., labels. Because these labels are available as ground truth, we can use them to precisely quantify registration accuracy through the Dice coefficient, avoiding inter- and intra-observer variabilities and other issues associated with establishing ground truth labels in real imaging data.
\end{itemize}

By performing image registration at different resolutions (and applying the resulting velocity to transform the high resolution original images), we want to check whether the registration at higher resolutions is more accurate than performing the registration at a lower resolution.

\paragraph{Dataset} We use the \textbf{SYN} dataset (see~\secref{s:datasets}) for this experiment.

\paragraph{Procedure} We execute registration at different resolutions for the original resolution images and quantify the accuracy using the Dice coefficient for labels before and after the registration. We compare the Dice statistics for different resolutions. More specifically, we take the following steps:
\begin{enumerate}
\item We register the template image $m_0$ to the reference image $m_1$ at the base resolution $n$ to get the velocity field $v_{\vect{n}}$. We transport $m_0$ using the velocity $v_{\vect{n}}$ to get the deformed template image $m(t=1)$ by solving~\eqref{e:varopt:constraint}. Then, we compute the Dice score between $l^i(t=1) $ and $ l_1^i $, $i \in {1,\ldots,10}$ which are discrete labels for $m(t=1)$ and $m_1$, respectively, using~\eqref{e:dice}.

\item We downsample $m_0$ and $m_1$ using nearest neighbor interpolation to half the base resolution (for example, $\vect{n}/2 = (512,512,512)$\footnote{We treat $\vect{n}_x = (N_1,N_2,N_3) $ as a tuple. When we say $ \vect{n}_x/2 $, we mean $ \vect{n}_x/2 = (N_1/2 , N_2/2 , N_3/2)$.} and register the downsampled images to get the velocity $\hat{v}_{\vect{n}/2}$. We upsample $\hat{v}_{\vect{n}/2}$ to the base resolution $\vect{n}$ using spectral prolongation and call it $v_{\vect{n}/2}$. We transport $m_0$ using $v_{\vect{n}/2}$ by solving~\eqref{e:varopt:constraint} to get the deformed template image $m(t=1)$ and then compute the Dice score for this new deformed template image.

\item We repeat the procedure in step 2 for resolutions $n/4$ and $n/8$ and compute the corresponding Dice scores.
\end{enumerate}

For the registration, we fix the determinant $J$ of the deformation gradient to be within $[5\text{e-}02, 20]$ and search for the regularization parameters using the proposed parameter search scheme as described above in \secref{s:methods}. Note that we perform a search for an optimal regularization parameter for each individual dataset because \textit{we want to obtain the best result for each pair of images}. In practical applications, this is not necessary (see comments below; we also refer to \cite{Mang:2018CLAIRE} for a discussion). We fix the tolerance for the reduction of the gradient to $5\text{e-}02$, which we have found to be sufficiently accurate for most image registration problems (see \cite{Mang:2018CLAIRE}). We use linear interpolation in the semi-Lagrangian scheme. Another hyperparameter in our registration solver is the number of time steps $n_t$ for the semi-Lagrangian (\acr{SL}) scheme. We consider two cases for selecting $n_t$:
\begin{enumerate}
\item \textbf{$n_t$ changes with resolution:} We use $n_t=4$ time steps for the coarsest resolution $\vect{n}_x=\vect{n}/8$ and double $n_t$ when we double the resolution in order to keep the CFL number fixed. All other solver parameters, except for the regularization parameters, are the same at each resolution.
\item \textbf{$n_t$ fixed with resolution:} In order to study the effect of $n_t$ on the Dice score we keep $n_t$ fixed for each $\vect{n}_x$, instead of increasing $n_t$ proportionately to $\vect{n}_x$.
\end{enumerate}

\begin{table}
\caption{\textbf{Experiment 2A: Registration performance for \claire{} for case 1 ($n_t$ changes proportionally to the image resolution, see~\secref{exp_1a})}. Comparison of registration accuracy based on the Dice score at different resolutions for the synthetic dataset \textbf{SYN}. $K$ denotes the frequency of the synthetic velocity field in~\eqref{e:syn_vel}. $n=(1024,1024,1024)$ is the base image resolution. We fix the tolerance for the reduction of the gradient to $5\text{e-}02$ and use linear interpolation. The Jacobian bounds for the parameter search are $[0.05,20]$. We report $\beta^\star_v$ and $\beta^\star_w$ (the optimal regularization parameters obtained with the proposed parameter search scheme), and $J_\text{min}$ and $J_\text{max}$ (the minimum and maximum values for the determinant of the deformation gradient). For the Dice score, we report average Dice ($D_a$), the volume weighted average Dice ($D_{vw}$), and the inverse volume weighted average Dice ($D_{ivw}$), pre and post registration. We also report the wall clock time for the parameter search.}\label{tab:syn_spherical_harmonic_reg_Dice_stats_iporder1}
\centering\scriptsize
\begin{tabular}{llllrrrrrRrRrRr}
\toprule
 run & K & $\vect{n}_x$ & $n_t$   & $\beta^*_v$ & $\beta^*_w$ & $J_{min}$ & $J_{max}$ &   \multicolumn{2}{c}{$D_a$}   & \multicolumn{2}{c}{$D_{vw}$}   & \multicolumn{2}{c}{$D_{ivw}$}          & runtime(s) \\
& & & & & & & & pre & post & pre & post & pre & post & search \\
\midrule
 \textbf{\#1} & \multirow{4}{*}{ 4} & $\vect{n}$   & 32 & 1.1e-05 & 1.0e-07 & 1.7e-01 & 7.4e+00 & \multirow{4}{*}{3.1e-01} & 9.2e-01 &  \multirow{4}{*}{5.8e-01} & 9.8e-01 &   \multirow{4}{*}{3.9e-02} & 8.5e-01 &    2.9e+03 \\
 \textbf{\#2} & & $\vect{n}$/2 & 16 & 1.1e-05 & 1.0e-07 & 1.9e-01 & 7.7e+00 & & 8.7e-01 & & 9.7e-01 & & 7.0e-01 & 6.5e+02 \\
 \textbf{\#3} & & $\vect{n}$/4 &  8 & 1.1e-05 & 1.0e-07 & 2.6e-01 & 1.4e+01 & & 7.9e-01 & & 9.5e-01 & & 5.0e-01 & 1.1e+02 \\
 \textbf{\#4} & & $\vect{n}$/8 &  4 & 1.1e-05 & 1.0e-06 & 4.7e-01 & 5.6e+00 & & 6.7e-01 & & 9.1e-01 & & 1.8e-01 & 1.5e+01 \\
\cline{1-15}
 \textbf{\#5} & \multirow{4}{*}{ 8} & $\vect{n}$ & 32 & 1.1e-05 & 1.0e-07 & 5.1e-02 & 1.0e+01 & \multirow{4}{*}{3.2e-01} & 9.0e-01 &  \multirow{4}{*}{5.3e-01} & 9.8e-01 &   \multirow{4}{*}{7.4e-02} & 7.6e-01 &    2.7e+03 \\
 \textbf{\#6} & & $\vect{n}$/2 & 16 & 1.1e-05 & 1.0e-07 & 1.8e-01 & 1.5e+01 & & 8.5e-01 & & 9.7e-01 & & 6.0e-01 & 6.2e+02 \\
 \textbf{\#7} & & $\vect{n}$/4 &  8 & 1.1e-05 & 1.0e-06 & 3.0e-01 & 7.8e+00 & & 7.6e-01 & & 9.4e-01 & & 4.1e-01 & 1.0e+02 \\
 \textbf{\#8} & & $\vect{n}$/8 &  4 & 2.4e-05 & 1.0e-06 & 3.8e-01 & 4.8e+00 & & 6.4e-01 & & 9.0e-01 & & 1.7e-01 & 1.4e+01 \\
\cline{1-15}
 \textbf{\#9} & \multirow{4}{*}{12} & $\vect{n}$   & 32 & 1.1e-05 & 1.0e-07 & 1.7e-01 & 1.2e+01 & \multirow{4}{*}{3.1e-01} & 9.2e-01 &  \multirow{4}{*}{5.2e-01} & 9.8e-01 &   \multirow{4}{*}{9.5e-02} & 8.5e-01 &    2.6e+03 \\
 \textbf{\#10} & & $\vect{n}$/2 & 16 & 1.1e-05 & 1.0e-06 & 3.1e-01 & 8.9e+00 & & 8.6e-01 & & 9.7e-01 & & 7.4e-01 & 5.4e+02 \\
 \textbf{\#11} & & $\vect{n}$/4 &  8 & 1.1e-05 & 1.0e-06 & 2.9e-01 & 1.2e+01 & & 7.5e-01 & & 9.4e-01 & & 4.5e-01 & 9.4e+01 \\
 \textbf{\#12} & & $\vect{n}$/8 &  4 & 1.1e-05 & 1.0e-06 & 4.1e-01 & 9.9e+00 & & 6.0e-01 & & 8.9e-01 & & 1.9e-01 & 1.4e+01 \\
\cline{1-15}
 \textbf{\#13} &\multirow{4}{*}{16} & $\vect{n}$   & 32 & 1.1e-05 & 1.0e-07 & 1.6e-01 & 9.5e+00 & \multirow{4}{*}{2.9e-01} & 9.1e-01 &  \multirow{4}{*}{5.1e-01} & 9.8e-01 &   \multirow{4}{*}{9.0e-02} & 8.1e-01 &    2.4e+03 \\
 \textbf{\#14} & & $\vect{n}$/2 & 16 & 1.1e-05 & 1.0e-07 & 1.7e-01 & 1.4e+01 &  & 8.4e-01 & & 9.7e-01 & & 6.0e-01 & 5.2e+02 \\
 \textbf{\#15} & & $\vect{n}$/4 &  8 & 1.4e-05 & 1.0e-06 & 3.0e-01 & 8.8e+00 &  & 7.4e-01 & & 9.4e-01 & & 4.7e-01 & 9.5e+01 \\
 \textbf{\#16} & & $\vect{n}$/8 &  4 & 2.7e-05 & 1.0e-06 & 3.9e-01 & 1.5e+01 &  & 6.1e-01 & & 9.0e-01 & & 2.0e-01 & 1.5e+01 \\
\bottomrule
\end{tabular}

\end{table}

\begin{table}
\caption{\textbf{Experiment 2A: Registration performance for \claire{} for case 2 ($n_t$ independent of the image resolution).} Comparison of registration accuracy using Dice at different resolutions for the synthetic dataset \textbf{SYN}. $K$ denotes the frequency of the synthetic velocity field in~\eqref{e:syn_vel}. $n=(1024,1024,1024)$ is the base image resolution. We fix the tolerance for the reduction of the gradient to $5\text{e-}02$ and use linear interpolation. The Jacobian bounds for parameter search is $[0.05,20]$. For each value of $n_t$, we report results for different resolutions. We report $\beta^\star_v$ and $\beta^\star_w$ (the optimal regularization parameters obtained with the proposed parameter search scheme), and $J_\text{min}$ and $J_\text{max}$ (the minimum and maximum values for the determinant of the deformation gradient). For the Dice score, we report average Dice ($D_a$), the volume weighted average Dice ($D_{vw}$), and the inverse volume weighted average Dice ($D_{ivw}$), pre and post the registration. We also report the wall clock time for the parameter search. The missing cases for $K=8$ failed to finish in a reasonable time frame. We only report a couple of cases for $K=16$ and expect a behavior similar to $K=8$ for the rest.}\label{tab:syn_spherical_harmonic_reg_Dice_stats_iporder1_nt_variation}
\centering\scriptsize

\begin{tabular}{llllrrrrrRrRrRr}
\toprule
run & K & $n_t$ & $\vect{n}_x$ & $\beta^\star_v$ & $\beta^\star_w$ & $J_\text{min}$ & $J_\text{max}$ & \multicolumn{2}{c}{$D_a$} & \multicolumn{2}{c}{$D_{vw}$} & \multicolumn{2}{c}{$D_{ivw}$} & runtime(s) \\
& & & & & & & & pre & post & pre & post & pre & post & search \\
\midrule

 \textbf{\#1}  & \multirow{12}{*}{8} & \multirow{3}{*}{4} & $\vect{n}$/2 & 1.4e-05 & 1.0e-07 & 9.7e-02 & 1.1e+01 & \multirow[c]{12}{*}{3.2e-01} & 8.8e-01 & \multirow[c]{12}{*}{5.3e-01} & 9.8e-01 & \multirow[c]{12}{*}{7.4e-02} & 7.4e-01 &    3.9e+02 \\
 \textbf{\#2}  & & & $\vect{n}$/4 & 1.1e-05 & 1.0e-07 & 3.8e-01 & 3.8e+00 & & 6.8e-01 & & 9.2e-01 & & 2.1e-01 & 7.9e+02 \\
 \textbf{\#3}  & & & $\vect{n}$/8 & 2.4e-05 & 1.0e-06 & 3.8e-01 & 4.8e+00 & & 6.2e-01 & & 8.9e-01 & & 1.6e-01 & 1.4e+01 \\
\cline{3-8}\cline{15-15}
 \textbf{\#4}  & &  \multirow{2}{*}{8} & $\vect{n}$/4 & 1.1e-05 & 1.0e-06 & 2.9e-01 & 7.7e+00 & & 7.5e-01 & & 9.4e-01 & & 4.1e-01 & 1.0e+02 \\
 \textbf{\#5}  & & & $\vect{n}$/8 & 1.7e-05 & 1.0e-06 & 3.9e-01 & 6.4e+00 & & 5.9e-01 & & 8.7e-01 & & 1.6e-01 & 1.6e+01 \\
\cline{3-8}\cline{15-15}
 \textbf{\#6}  & & \multirow{3}{*}{16}  & $\vect{n}$/2 & 1.1e-05 & 1.0e-07 & 1.8e-01 & 1.4e+01 & & 8.3e-01 & & 9.7e-01 & & 5.2e-01 & 5.9e+02 \\
 \textbf{\#7}  & & & $\vect{n}$/4 & 1.1e-05 & 1.0e-06 & 3.1e-01 & 8.2e+00 & & 7.1e-01 & & 9.2e-01 & & 3.4e-01 & 1.2e+02 \\
 \textbf{\#8}  & & & $\vect{n}$/8 & 1.1e-05 & 1.0e-07 & 5.4e-01 & 3.2e+00 & & 5.6e-01 & & 8.5e-01 & & 1.4e-01 & 6.7e+01 \\
 \cline{3-8}\cline{15-15}
 \textbf{\#9}  & & \multirow{4}{*}{32}& $\vect{n}$ & 1.1e-05 & 1.0e-07 & 5.1e-02 & 1.0e+01 & & 9.0e-01 & & 9.8e-01 & & 7.6e-01 & 2.7e+03 \\
 \textbf{\#10} & & & $\vect{n}$/2 & 1.1e-05 & 1.0e-07 & 1.2e-01 & 1.9e+01 & & 7.8e-01 & & 9.5e-01 & & 4.2e-01 & 7.6e+02 \\
 \textbf{\#11} & & & $\vect{n}$/4 & 1.1e-05 & 1.0e-06 & 3.1e-01 & 1.0e+01 & & 6.8e-01 & & 9.0e-01 & & 3.3e-01 & 1.9e+02 \\
 \textbf{\#12} & & & $\vect{n}$/8 & 1.1e-05 & 1.0e-07 & 5.2e-01 & 3.2e+00 & & 5.6e-01 & & 8.5e-01 & & 1.4e-01 & 4.8e+01 \\
 \cline{2-15}
 \textbf{\#13} &\multirow{2}{*}{16} &  4 & $n$/2 & 1.3e-05 & 1.0e-06 & 2.0e-01 & 6.9e+00 & \multirow[c]{2}{*}{2.9e-01} & 8.6e-01 & \multirow[c]{2}{*}{5.1e-01} & 9.7e-01 & \multirow[c]{2}{*}{9.0e-02} & 7.8e-01 & 3.7e+02 \\
 \textbf{\#14} & & 32 & $\vect{n}$ & 1.1e-05 & 1.0e-07 & 1.6e-01 & 9.5e+00 & & 9.1e-01 & & 9.8e-01 & & 8.1e-01 & 2.4e+03 \\
\bottomrule
\end{tabular}
\end{table}

\begin{figure}
\centering
\includegraphics[width=0.55\linewidth]{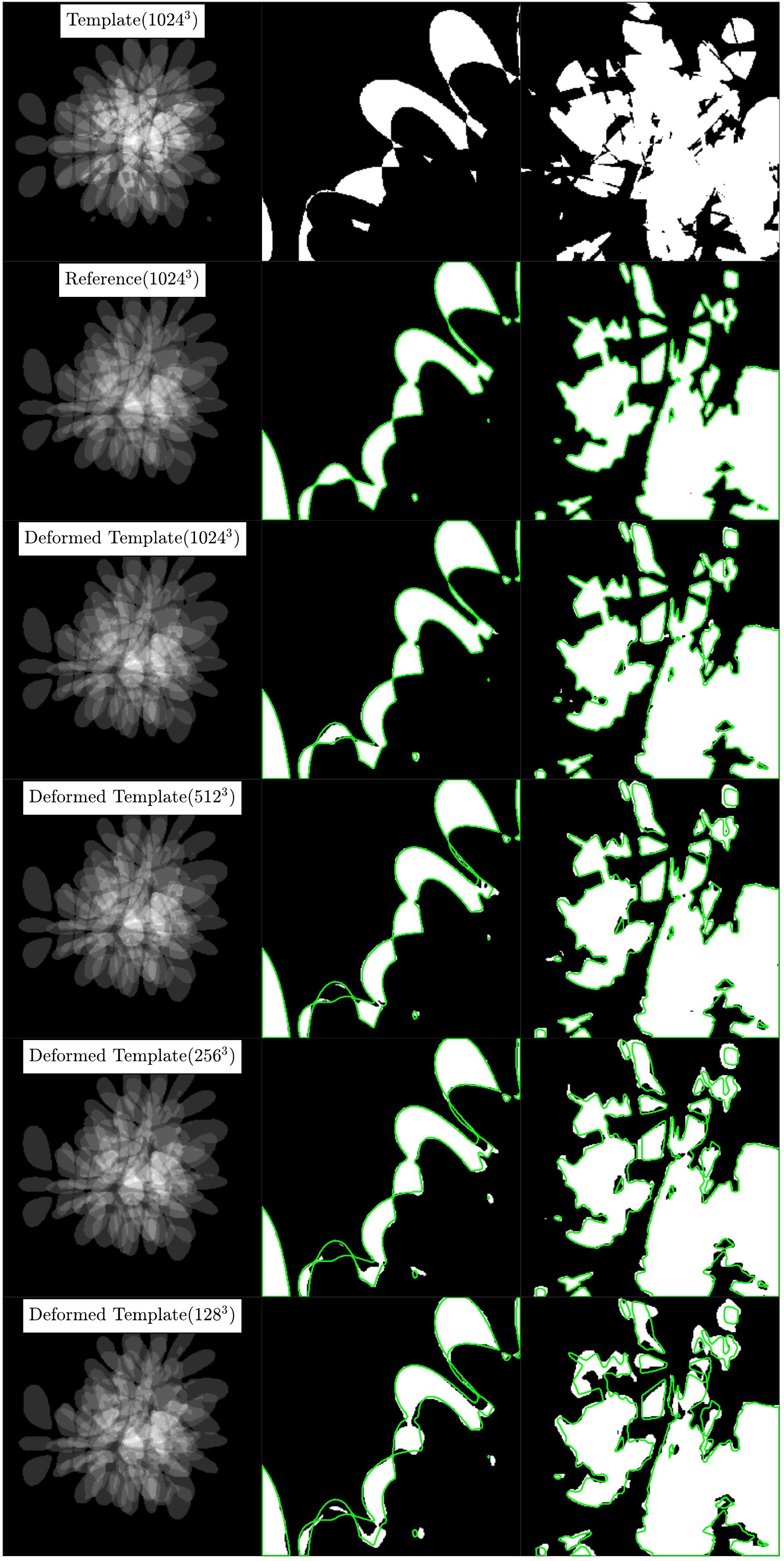}
\caption{\textbf{Experiment 2A: Visualization of registration results for case 1.} In column 1, from top to bottom, we visualize the template, reference and deformed template images for registrations done at different resolutions. These images correspond to the runs~\#1-4 in~\tabref{tab:syn_spherical_harmonic_reg_Dice_stats_iporder1}. The value in the parentheses in column 1 indicates the resolution at which registration was done. The visualization is done at the original resolution $\vect{n} = (1024,1024,1024)$. In column 2 and 3, we visualize cropped portions of the images shown in column 1 for specific label values. In column 2, we show label 1, in column 3, we show the union of labels with intensity value $\geq 5$. Note that higher label values have smaller volumes and more fine-grained features. We plot the label boundaries for the reference image in green to visualize the registration errors.}\label{fig:synthetic-spherical-harmonics-K-8_sl2}
\end{figure}

\begin{figure}
\centering
\includegraphics[width=0.7\linewidth]{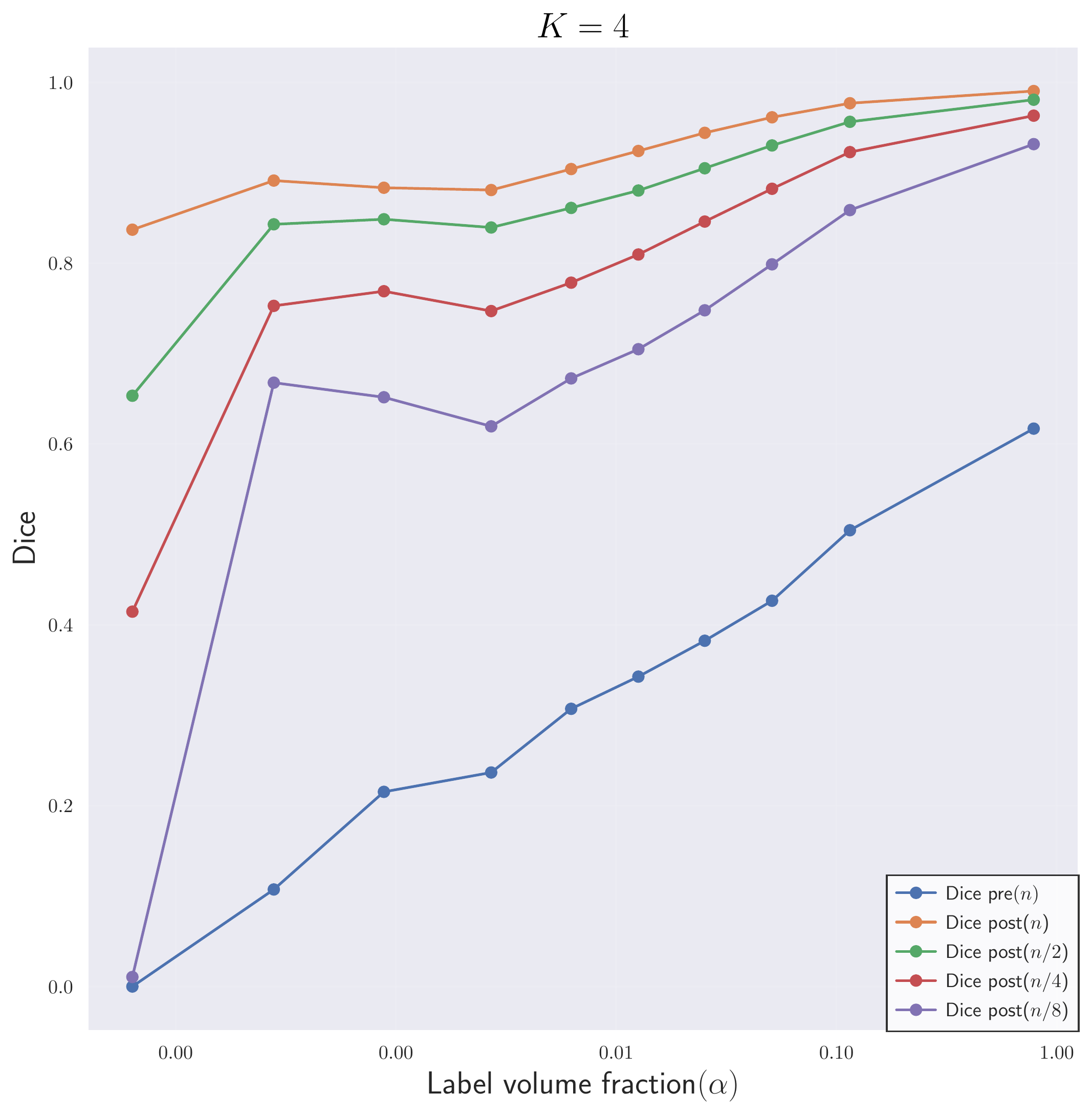}
\caption{\textbf{Experiment 2A: Quantitative results for the registration results corresponding to case 1}. We show a plot of the Dice scores against the label volume fraction $\alpha$ for each label $l^i$, $i=1,\ldots,10$ for the registration of the synthetic data set \textbf{SYN} at different resolutions. This figure corresponds to the registration runs \#1-4 in~\tabref{tab:syn_spherical_harmonic_reg_Dice_stats_iporder1} for $K=4$.}\label{fig:synthetic_dice_vol_plots}
\end{figure}

\begin{figure}
\centering
\includegraphics[width=1\linewidth]{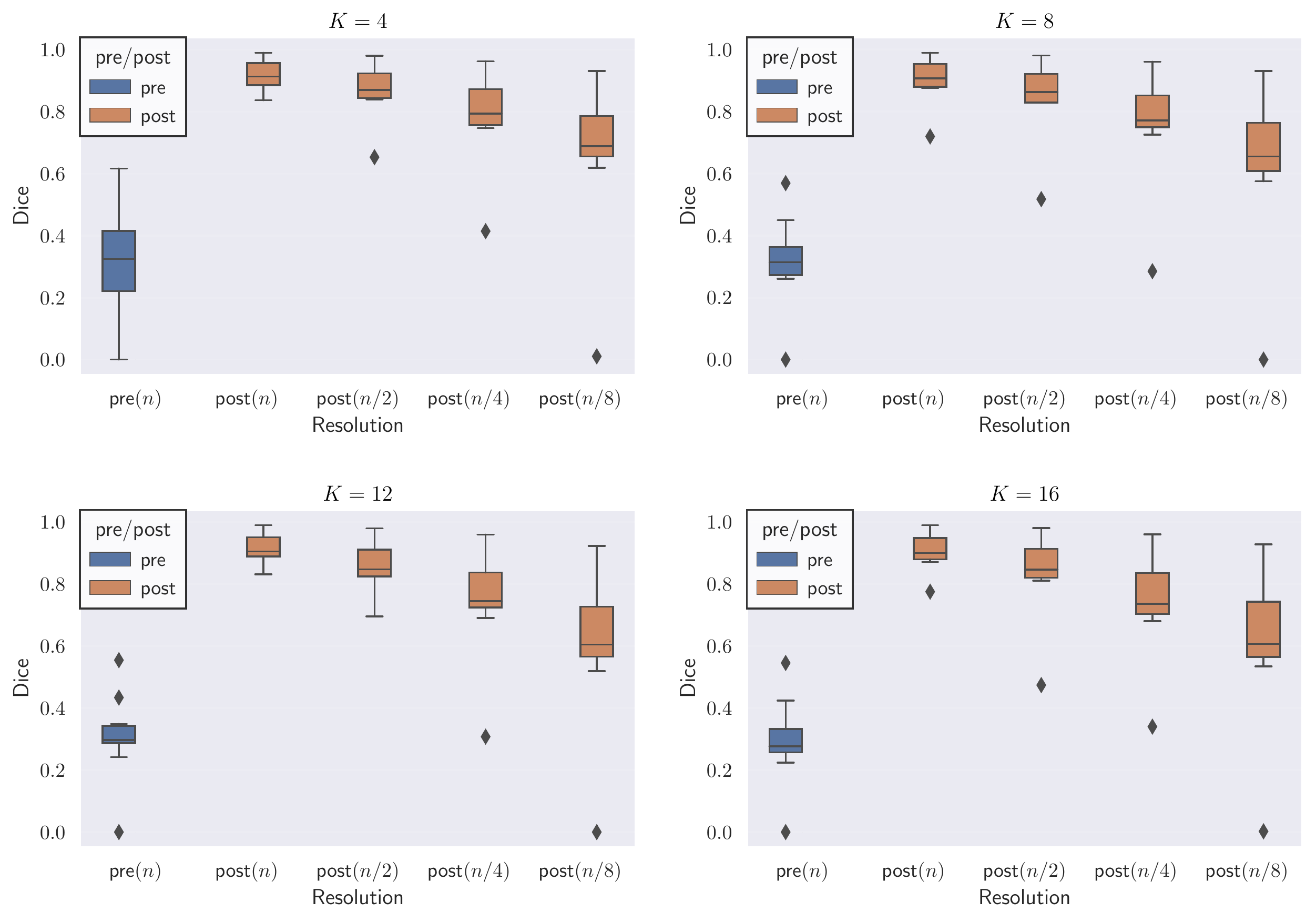}
\caption{\textbf{Experiment 2A: Quantitative results for the registration results corresponding to case 1.} We show box plots of the Dice scores for the individual labels before and after registration for different resolutions. We consider the synthetic test problem \textbf{SYN}. This figure corresponds to the registration results reported in~\tabref{tab:syn_spherical_harmonic_reg_Dice_stats_iporder1}.}\label{fig:synthetic_dice_box_plots}
\end{figure}

\paragraph{Results} In~\figref{fig:synthetic-spherical-harmonics-K-8_sl2}, we visualize the template, reference and deformed template images for the synthetic problem constructed with $K=4$.  We report quantitative results for \claire{} in~\tabref{tab:syn_spherical_harmonic_reg_Dice_stats_iporder1} and~\tabref{tab:syn_spherical_harmonic_reg_Dice_stats_iporder1_nt_variation}, respectively. In~\figref{fig:synthetic_dice_vol_plots}, we compare the Dice score for individual labels as a function of their volume fraction $\alpha$. In~\figref{fig:synthetic_dice_box_plots}, we visualize box plots of the Dice score for the registrations reported in~\tabref{tab:syn_spherical_harmonic_reg_Dice_stats_iporder1}.

\paragraph{Observations} The most important observations are: \emph{(i)} The Dice score averages are better for registrations performed at the base resolution $\vect{n}$ with progressively worse Dice scores for registrations done at coarser resolutions. \emph{(ii)} The difference between Dice scores for registrations done at successively coarser resolutions for $K=16$ (rougher velocity field) is higher than at $K=4$ (smoother velocity field). \emph{(iii)} Keeping $n_t$ fixed for the base and coarser resolutions does not affect the Dice score trend, i.e., the Dice decreases as $\vect{n}_x$ is decreased. In the following, we give more details for these general observations.

Regarding Dice score averages in~\tabref{tab:syn_spherical_harmonic_reg_Dice_stats_iporder1}, we observe that $D_a$, the arithmetic mean of the Dice scores of individual labels, drops by as much as 7\% between run \#13 and \#14. However, the percentage drop in volume weighted Dice average $D_{vw}$ is smaller than in $D_a$. This indicates that labels with higher volume are still easier to register at coarser resolutions. The inverse weighted Dice average $D_{ivw}$, which gives more weight to smaller labels, features a more pronounced decrease because smaller regions contribute to the high frequency content in the image; this information is lost when the images are downsampled. We observe a 21.8\% difference in $D_{ivw}$ for the high frequency images in run \#13 and \#14 for $K=16$. As we increase the frequency $K$ of the synthetic ground truth velocity, we see that the difference in all Dice score averages between successive resolutions increases. As $K$ increases, we get increasingly rougher velocity fields, which we can not recover by registering the original images at coarser resolutions.

In~\tabref{tab:syn_spherical_harmonic_reg_Dice_stats_iporder1_nt_variation}, the Dice scores behave the same way even when $n_t$ is fixed for different $\vect{n}_x$, indicating that the loss in accuracy is primarily because of the reduction in the spatial resolution (and not the temporal resolution). We also observe that for the full resolution of $\vect{n}_x=\vect{n}$, using $n_t < 32$ results in slow solver convergence; the run did not finish in under 2~hrs. We attribute this slow convergence rate to the loss in numerical accuracy in the computation of the reduced gradient in~\eqref{e:reducedgrad}. If we compare run \#1 and \#9 in \tabref{tab:syn_spherical_harmonic_reg_Dice_stats_iporder1_nt_variation}, we see that the difference in $D_a$ is marginal in comparison to the run time cost overhead for run \#9. However, the accuracy difference increases as $K$ is increased, and the images get less smooth (see runs \#13 and \#14).

These quantitative observations are confirmed by the visual analysis in the figures shown: From~\figref{fig:synthetic-spherical-harmonics-K-8_sl2}, we observe that at lower resolutions (top to bottom), the alignment of the outlines (green lines; reference image) with the structures (white areas; deformed template image) is less accurate. \figref{fig:synthetic_dice_vol_plots}: shows that the Dice score is worse for labels with smaller volume fractions, i.e., fine structures are matched less accurately at coarse resolutions. Looking at~\figref{fig:synthetic_dice_box_plots}, we observe that the average registration accuracy decreases as we decrease the resolution.

We use 32 GPUs for registration at $\vect{n}_x = (1024,1024,1024)$ 4 GPUs for $\vect{n}_x = (512,512,512)$ and a single GPU for $\vect{n}_x = (256,256,256)$ and $\vect{n}_x = (128,128,128)$. Registration for $\vect{n}_x = (1024,1024,1024)$ takes on average 44 minutes wall clock time. It is important to note that this includes the time spent in the search for optimal regularization parameters (i.e., we solve the inverse problem multiple times using warm starts; see \secref{s:param_cont} for details regarding the scheme). For the large-scale runs that use multiple GPUs, the overall runtime of the solver is dominated by communication between MPI processes~\cite{Brunn:2020a}. Adding more resources does not necessarily reduce the runtime because of this increase in communication cost. Registrations for $\vect{n}_x = (512,512,1512)$ and lower resolutions are much quicker and run in the order of 10 min or less. In the present work, we perform the parameter search for each individual case because \textit{we want to obtain the best result for each pair of images}. However, in practice where a medical imaging pipeline requires registrations for several similar images, we suggest running the parameter search scheme on one pair of images and use the obtained regularization parameters to run the cohort registration for all images, as we have done in our previous work~\cite{Mang:2018CLAIRE}. This strategy reduces the computational cost drastically. One downside to this strategy is that some images in the cohort will not be registered as accurately as others.

Our experiment with synthetic images suggests that Dice scores are better when registrations are done in the original, high resolution at which the labels were created. Registration accuracy is affected more significantly if high frequency velocity fields are considered. The images used in this experiment are synthetic and free of noise. We use these images for both registration and evaluation of performance using Dice scores. Because the ground truth labels for these images at the highest resolution are known with certainty, we have high confidence in our observations regarding registration accuracy: the Dice scores become worse when registration is conducted at lower resolutions. However, in practical applications, images have noise and low contrast. To evaluate the registration accuracy for real images using Dice scores, we first evaluate their segmentation using external segmentation tools. This segmentation step is prone to errors (not only due to noise and a lack of contrast but also due to inherent limitations in segmentation software themselves). These errors result in a misalignment between the structures present in the original image and its segmentation, which complicates our analysis. Having said this, we conduct experiments on real brain MRIs in the next section to explore if we can provide experimental evidence that at least partially confirms the observations we have made in this section.

\subsection{Experiment 2B: High Resolution Real Data Registration}\label{exp_1b}

In this experiment, we aim at answering \textbf{Q1} as well as \textbf{Q2}. We do this by registering real human brain MRI datasets instead of synthetic images. Unlike synthetic images, these images are not noise-free. Moreover, they lack high contrast.

\paragraph{Datasets} We use the \textbf{NIREP} and the \textbf{MRI250} image datasets (see~\secref{s:datasets}) for this experiment.

\paragraph{Procedure} We designate the MRI250 image as the template image $m_0$. We generate the reference images $m_1$ from the images \texttt{na01}--\texttt{na10} from the NIREP dataset since we do not have access to other $T_1$-weighted MRI from a different subject at the original resolution of 250~$\mu$m. The acquired spatial resolution of the NIREP data is 1~mm, which is $4\times$ larger than 250~$\mu$m. Therefore, in order to generate a reference image $m_1$ that are 250~$\mu$m in spatial resolution, we take the following steps:
\begin{enumerate}
\item Upsample the respective NIREP image from $256\times300\times256$ to $640\times880\times880$ using linear interpolation.
\item Register MRI250 to the upsampled NIREP image using \claire{} and transport $m_0$ (which corresponds to the MRI250 image) using the resulting velocity $\vect{v}$ and solving~\eqref{e:varopt:constraint} to obtain the deformed template image $m_1 = m(t=1)$. We set the tolerance for the relative gradient norm to $g_\text{tol} = 1\text{e-}02$. We lower the tolerance compared to other runs to obtain a potentially more accurate registration result. We use the default regularization parameters $\beta_v=1\text{e-}02$ and $\beta_w=1\text{e-}04$. Consequently, we do not perform a parameter search to estimate an optimal regularization parameter for this registration. We want to keep the downstream registration performance analysis, where we will use parameter search, oblivious to the process of generating the high resolution reference image.
\end{enumerate}

To generate a segmentation that we can use to compute Dice scores (not for the registration itself, which is done on the original unsegmented images), we use the tool \texttt{fast} from FSL~\cite{jenkinsonFSL2012a} both on the template image $m_0$ and on the reference image $m_1$. We generate labels \emph{WM, GM}, and \emph{CSF}. The remaining steps for this experiment are the same as described in experiment 2A in~\secref{exp_1a} except that here we are registering real $T_1$-weighted images instead of noise-free synthetic images. The base resolution for this experiment is $\vect{n}_x = \vect{n} =  (640,880,880)$. We consider $\vect{n}_x=\vect{n}/2$ and $\vect{n}_x=\vect{n}/4$ for the downsampled resolutions. We also consider the two sub-cases for selecting $n_t$ as we did in~\secref{exp_1a}. For the case where $n_t$ changes with resolution, we use $n_t=4$ for $\vect{n}_x=n/4$, $n_t=8$ for $\vect{n}_x=n/2$ and $n_t=16$ for $\vect{n}_x=n$.

\begin{figure}
\centering
\includegraphics[width=\linewidth]{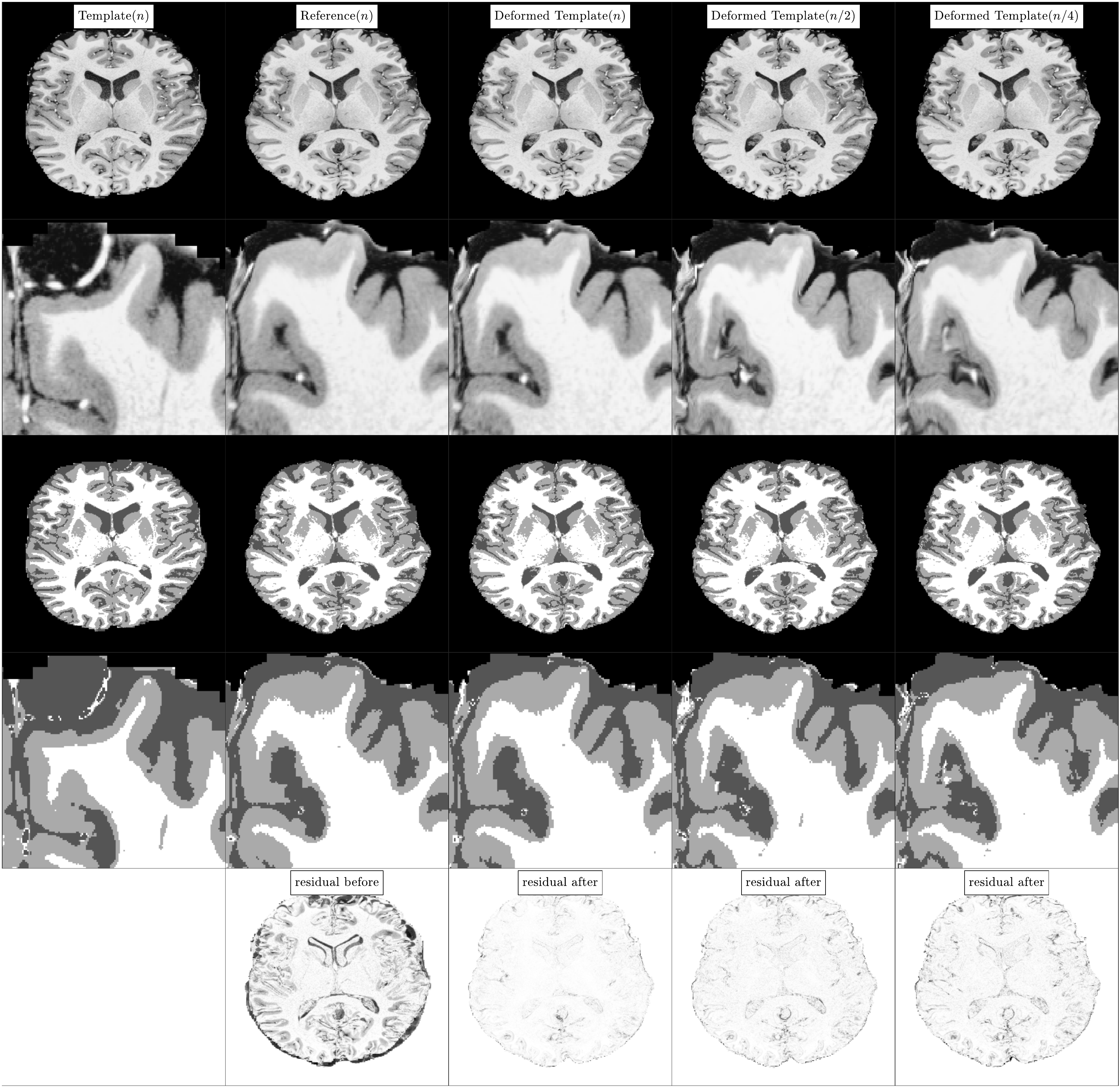}
\caption{\textbf{Experiment 2B: Illustration of registration results for the multi-resolution registration experiment on real brain images.} The images shown here correspond to the runs \#1, \#2, and \#3 in~\tabref{tab:brainMRI250um_nirep_velocity_multilevel_Dice_stats_iporder1}). The base resolution is $\vect{n}_x=\vect{n}=(640,880,880)$. In row 1, from left to right, we show the T1-weighted MRI250 datasets (template image $m_0$), the upsampled \texttt{na01} dataset (reference image $m_1$) from the NIREP data repository, and the deformed template images obtained from registration at resolutions $\vect{n}_x$, $\vect{n}_x/2$ and $\vect{n}_x/4$, respectively. In row 2, we show a cropped portion of the images from row 1. In rows 3 and 4, we show the label maps consisting of white matter (WM; white), gray matter (GM; light gray) and cerebro-spinal fluid (CSF; dark gray) and their cropped versions, respectively. In row 5, we show the image residuals before and after registration with respect to each resolution level.}\label{fig:exp_1b}
\end{figure}

\begin{table}
\centering\small
\caption{\textbf{Experiment 2B: Registration performance for \claire{} for case 1 ($n_t$ changes proportional to the image resolution)}. Comparison of registration accuracy using Dice and relative residual $r$ at different resolutions for the registration of the MRI250 brain image to templates generated from ten \textit{real} MRI scans from the NIREP dataset. We consider three resolution levels $\vect{n}_x=\{\vect{n}, \vect{n}/2, \vect{n}/4\}$ where $\vect{n} = (640,880,880)$. We fix the tolerance for the relative gradient to $5\text{e-}02$. We use linear interpolation in the semi Lagrangian scheme. The bounds for the determinant of the deformation gradient for the parameter search are $[0.05,20]$.  We report the regularization parameters $\beta_v^\star$ and $\beta_w^\star$ obtained through the proposed parameter search scheme, the minimum and maximum determinant of the deformation gradient ($J_\text{min}$ and $J_\text{max}$), the relative residual ($r$), the average Dice ($D_a$), pre and post the registration, as well as the wall clock time for the parameter search.}\label{tab:brainMRI250um_nirep_velocity_multilevel_Dice_stats_iporder1}
\begin{tabular}{llllrrrrrrRr}
\toprule
run & NIREP & $\vect{n}_x$ & $n_t$  & $\beta^*_v$ & $\beta^*_w$ & $J_{min}$ & $J_{max}$ &     $r$ &   \multicolumn{2}{c}{$D_a$}        & runtime(s) \\
     &     &    &               &           &           &           &           &         &     pre &    post &     search       \\
\midrule
\textbf{\#1} & \multirow{3}{*}{na01} & $\vect{n}$ & 16 & 1.1e-03 & 1.0e-05 & 1.8e-01 & 8.3e+00 & 2.5e-01 & \multirow{3}{*}{5.5e-01} & 9.0e-01 & 3.1e+02 \\
\textbf{\#2} &  & $\vect{n}$/2 & 8 & 1.1e-05 & 1.0e-06 & 1.8e-01 & 6.5e+00 & 3.5e-01 & & 8.5e-01 & 3.4e+02 \\
\textbf{\#3} &  & $\vect{n}$/4 & 4 & 1.1e-05 & 1.0e-05 & 2.3e-01 & 8.2e+00 & 4.5e-01 & & 8.0e-01 & 3.6e+01 \\
\cline{2-12}
\textbf{\#4} & \multirow{3}{*}{na02} & $\vect{n}$ & 16 & 1.1e-03 & 1.0e-05 & 7.8e-02 & 6.3e+00 & 2.5e-01 & \multirow{3}{*}{5.4e-01} & 8.9e-01 & 3.3e+02 \\
\textbf{\#5} & & $\vect{n}$/2 & 8 & 1.1e-05 & 1.0e-06 & 1.2e-01 & 4.3e+00 & 3.6e-01 & & 8.3e-01 & 3.0e+02 \\
\textbf{\#6} & & $\vect{n}$/4 & 4 & 1.1e-05 & 1.0e-06 & 8.1e-02 & 1.1e+01 & 4.6e-01 & & 7.7e-01 & 4.0e+01 \\
\cline{2-12}
\textbf{\#7} & \multirow{3}{*}{na03} & $\vect{n}$ & 16 & 1.1e-05 & 1.0e-07 & 1.1e-01 & 6.1e+00 & 3.3e-01 & \multirow{3}{*}{5.1e-01} & 8.4e-01 & 3.2e+03 \\
\textbf{\#8} & & $\vect{n}$/2 & 8 & 1.1e-05 & 1.0e-06 & 1.1e-01 & 1.8e+01 & 3.9e-01 & & 8.0e-01 & 2.9e+02 \\
\textbf{\#9} & & $\vect{n}$/4 & 4 & 1.1e-05 & 1.0e-07 & 1.0e-01 & 1.7e+01 & 4.7e-01 & & 7.6e-01 & 4.2e+01 \\
\cline{2-12}
\textbf{\#10} & \multirow{3}{*}{na04} & $\vect{n}$ & 16 & 3.1e-02 & 1.0e-05 & 1.2e-01 & 1.4e+01 & 3.7e-01 & \multirow{3}{*}{5.3e-01} & 8.0e-01 & 1.9e+02 \\
\textbf{\#11} & & $\vect{n}$/2 & 8 & 1.1e-03 & 1.0e-05 & 6.8e-02 & 8.9e+00 & 2.9e-01 & & 8.7e-01 & 5.1e+01 \\
\textbf{\#12} & & $\vect{n}$/4 & 4 & 1.1e-05 & 1.0e-05 & 1.0e-01 & 6.8e+00 & 4.6e-01 & & 7.6e-01 & 3.7e+01 \\
\cline{2-12}
\textbf{\#13} & \multirow{3}{*}{na05} & $\vect{n}$ & 16 & 1.1e-05 & 1.0e-05 & 9.5e-02 & 8.9e+00 & 3.2e-01 & \multirow{3}{*}{5.3e-01} & 8.5e-01 & 2.8e+03 \\
\textbf{\#14} & & $\vect{n}$/2 & 8 & 1.1e-05 & 1.0e-05 & 1.6e-01 & 1.1e+01 & 3.6e-01 & & 8.3e-01 & 2.5e+02 \\
\textbf{\#15} & & $\vect{n}$/4 & 4 & 1.1e-05 & 1.0e-05 & 1.6e-01 & 1.6e+01 & 4.5e-01 & & 7.8e-01 & 3.7e+01 \\
\cline{2-12}
\textbf{\#16} & \multirow{3}{*}{na06} & $\vect{n}$ & 16 & 1.1e-03 & 1.0e-05 & 7.8e-02 & 1.4e+01 & 2.5e-01 & \multirow{3}{*}{5.3e-01} & 8.9e-01 & 3.3e+02 \\
\textbf{\#17} & & $\vect{n}$/2 & 8 & 1.1e-05 & 1.0e-06 & 2.0e-01 & 5.6e+00 & 3.5e-01 & & 8.3e-01 & 3.0e+02 \\
\textbf{\#18} & & $\vect{n}$/4 & 4 & 1.1e-05 & 1.0e-05 & 1.6e-01 & 7.2e+00 & 4.4e-01 & & 7.7e-01 & 3.6e+01 \\
\cline{2-12}
\textbf{\#19} & \multirow{3}{*}{na07} & $n$ & 16 & 1.0e-02 & 1.0e-05 & 9.5e-02 & 2.0e+01 & 3.0e-01 & \multirow{3}{*}{5.3e-01} & 8.6e-01 & 2.4e+02 \\
\textbf{\#20} & & $\vect{n}$/2 & 8 & 1.1e-05 & 1.0e-05 & 1.6e-01 & 1.6e+01 & 3.5e-01 & & 8.4e-01 & 3.9e+02 \\
\textbf{\#21} & & $\vect{n}$/4 & 4 & 1.1e-05 & 1.0e-05 & 1.7e-01 & 1.7e+01 & 4.5e-01 & & 7.7e-01 & 3.7e+01 \\
\cline{2-12}
\textbf{\#22} & \multirow{3}{*}{na08} & $\vect{n}$ & 16 & 1.1e-05 & 1.0e-07 & 1.3e-01 & 4.8e+00 & 3.1e-01 & \multirow{3}{*}{5.3e-01} & 8.6e-01 & 2.5e+03 \\
\textbf{\#23} & & $\vect{n}$/2 & 8 & 1.1e-05 & 1.0e-06 & 1.0e-01 & 1.3e+01 & 3.8e-01 & & 8.1e-01 & 3.0e+02 \\
\textbf{\#24} & & $\vect{n}$/4 & 4 & 1.1e-05 & 1.0e-06 & 9.4e-02 & 1.7e+01 & 4.7e-01 & & 7.5e-01 & 4.2e+01 \\
\cline{2-12}
\textbf{\#25} & \multirow{3}{*}{na09} & $\vect{n}$ & 16 & 1.1e-03 & 1.0e-05 & 6.3e-02 & 1.5e+01 & 2.5e-01 & \multirow{3}{*}{5.3e-01} & 8.9e-01 & 3.5e+02 \\
\textbf{\#26} & & $\vect{n}$/2 & 8 & 1.1e-05 & 1.0e-05 & 1.2e-01 & 5.1e+00 & 3.5e-01 & & 8.3e-01 & 2.3e+02 \\
\textbf{\#27} & & $\vect{n}$/4 & 4 & 1.1e-05 & 1.0e-06 & 9.9e-02 & 7.4e+00 & 4.5e-01 & & 7.6e-01 & 4.2e+01 \\
\cline{2-12}
\textbf{\#28} & \multirow{3}{*}{na10} & $\vect{n}$ & 16 & 1.1e-05 & 1.0e-07 & 1.1e-01 & 5.7e+00 & 3.2e-01 & \multirow{3}{*}{5.4e-01} & 8.5e-01 & 2.6e+03 \\
\textbf{\#29} & & $\vect{n}$/2 & 8 & 1.1e-05 & 1.0e-05 & 1.2e-01 & 4.5e+00 & 3.5e-01 & & 8.3e-01 & 2.7e+02 \\
\textbf{\#30} & & $\vect{n}$/4 & 4 & 1.1e-05 & 1.0e-06 & 1.0e-01 & 9.1e+00 & 4.7e-01 & & 7.6e-01 & 4.1e+01 \\
\bottomrule
\end{tabular}
\end{table}

\begin{table}
\centering\small
\caption{\textbf{Experiment 2B: Registration performance for \claire{} for case 2 ($n_t$ independent of the image resolution)}. Comparison of registration accuracy using Dice and relative residual $r$ for a fixed number of time steps $n_t$ at different resolutions for the registration of the real MRI datasets MRI250 and the reference image $m_1$ generated from \texttt{na01} from the NIREP repository. We consider three resolution levels $\vect{n}_x=\{\vect{n},\vect{n}/2,\vect{n}/4\}$ where $n=(640,880,880)$. We fix the tolerance for the relative gradient to $5\text{e-}02$. We use linear interpolation in the semi Lagrangian schreme. The bounds for the determinant of the deformation gradient for the parameter search are $[0.05,20]$. We keep the time step $n_t$ fixed. We report the regularization parameters $\beta_v^\star$ and $\beta_w^\star$ obtained through the proposed parameter search scheme, the minimum and maximum determinant of the deformation gradient ($J_\text{min}$ and $J_\text{max}$), the relative residual ($r$), the average Dice ($D_a$), pre and post the registration, as well as the wall clock time for the parameter search. The case with $\vect{n}_x=\vect{n}$ and $n_t=4$ failed to finish in under 4~hrs.}\label{tab:brainMRI250um_nirep_velocity_multilevel_Dice_stats_iporder1_nt_variation}
\begin{tabular}{llllrrrrrrRr}
\toprule
run & NIREP & $n_t$ & $\vect{n}_x$ & $\beta^\star_v$ & $\beta^\star_w$ & $J_\text{min}$ & $J_\text{max}$ & $r$ & \multicolumn{2}{c}{$D_a$} & runtime(s) \\
& & & & & & & & & pre & post & search \\
\midrule
 \textbf{\#1} & \multirow{8}{*}{na01} & \multirow{2}{*}{ 4} & $\vect{n}$/2 & 1.1e-05 & 1.0e-06 & 9.2e-02 & 5.2e+00 & 3.1e-01 & \multirow{8}{*}{5.5e-01} & 8.7e-01 & 2.7e+02 \\
 \textbf{\#2} &                       &                     & $\vect{n}$/4 & 1.1e-05 & 1.0e-05 & 2.3e-01 & 8.2e+00 & 4.5e-01 &  & 8.0e-01 & 3.6e+01 \\
\cline{3-9}\cline{11-12}
 \textbf{\#3} &                       & \multirow{3}{*}{ 8} & $\vect{n}$   & 5.6e-03 & 1.0e-05 & 1.1e-01 & 1.1e+01 & 2.4e-01 &  & 9.0e-01 & 2.6e+02 \\
 \textbf{\#4} &                       &                     & $\vect{n}$/2 & 1.1e-05 & 1.0e-06 & 1.9e-01 & 6.6e+00 & 3.5e-01 &  & 8.5e-01 & 3.1e+02 \\
 \textbf{\#5} &                       &                     & $\vect{n}$/4 & 1.1e-05 & 1.0e-05 & 2.7e-01 & 1.2e+01 & 4.7e-01 &  & 7.8e-01 & 4.3e+01 \\
\cline{3-9}\cline{11-12}
 \textbf{\#6} &                       & \multirow{3}{*}{16} & $\vect{n}$   & 1.1e-03 & 1.0e-05 & 1.8e-01 & 8.4e+00 & 2.5e-01 &  & 9.0e-01 & 3.2e+02 \\
 \textbf{\#7} &                       &                     & $\vect{n}$/2 & 1.1e-05 & 1.0e-05 & 2.6e-01 & 7.6e+00 & 3.8e-01 &  & 8.3e-01 & 3.6e+02 \\
 \textbf{\#8} &                       &                     & $\vect{n}$/4 & 1.1e-05 & 1.0e-05 & 2.8e-01 & 1.6e+01 & 4.8e-01 &  & 7.7e-01 & 5.5e+01 \\
\bottomrule
\end{tabular}
\end{table}

\paragraph{Results} We report the solver parameters for our registration with \claire{} along with the relative residual $r$ and Dice score averages for \emph{GM}, \emph{WM} and \emph{CSF} before and after the registration in~\tabref{tab:brainMRI250um_nirep_velocity_multilevel_Dice_stats_iporder1}. The relative residual $r$ and the Dice score are always computed at the base resolution $\vect{n}=(640,880,880)$. The respective results with $n_t$ fixed independent of the resolution are given in~\tabref{tab:brainMRI250um_nirep_velocity_multilevel_Dice_stats_iporder1_nt_variation} for \texttt{na01}. We visualize the image registration results for the reference image \texttt{na01} in~\figref{fig:exp_1b}.

\paragraph{Observations} The most important observation is that the relative residual $r$ increases and Dice score averages decrease for registrations done at coarser resolutions irrespective of whether we increase $n_t$ proportionally to the resolution, see~\tabref{tab:brainMRI250um_nirep_velocity_multilevel_Dice_stats_iporder1} or keep $n_t$ fixed for different $\vect{n}_x$, see~\tabref{tab:brainMRI250um_nirep_velocity_multilevel_Dice_stats_iporder1_nt_variation}. This observation is in line with the experiment for the synthetic dataset \textbf{SYN} in~\secref{exp_1a}. Except for the case of \texttt{na04} (see runs \#10 and \#11 in \tabref{tab:brainMRI250um_nirep_velocity_multilevel_Dice_stats_iporder1}), all other cases exhibit increasingly worse registration performance at coarser resolutions.

In~\ref{exp_1a}, we used synthetic, noise-free, high-contrast images for assessing the registration accuracy at different resolutions. Here, we repeat the same experiment with real world images---T1-weighted MR images of the human brain. We used an external software to segment these images to provide the necessary labels to be able to quantify registration performance in terms of Dice score. Notice that this additional segmentation step will inevitably introduce additional errors to our analysis. Due to these additional errors at the native resolution, we expect that the improvement in registration performance at high resolution may not be as pronounced as for the synthetic images considered in~\ref{exp_1a} (which did not require this additional segmentation step). This hypothesis is confirmed if we compare the average Dice score $D_a$ across experiments.  In particular, if we reduce the resolution from $\vect{n}$ to $\vect{n}/4$ in~\ref{exp_1a} (see~\tabref{tab:syn_spherical_harmonic_reg_Dice_stats_iporder1}) and~\ref{exp_1b} (see~\tabref{tab:brainMRI250um_nirep_velocity_multilevel_Dice_stats_iporder1}), the Dice score drops by 15.25\% compared to 9.5\%, respectively.

In~\tabref{tab:brainMRI250um_nirep_velocity_multilevel_Dice_stats_iporder1_nt_variation}, the case with $n_t=4$ and $\vect{n}_x=\vect{n}$ took very long to converge (>4 hrs). For this case the $CFL$ number is $15.66$ during the inverse solve while for $n_t=16$, the $CFL$ number is $4$. The larger $CFL$ number for $n_t=4$ yields a higher adjoint error in the SL scheme. This leads to higher errors in the computation of the reduced gradient, which results in worse convergence of the inverse solver for $n_t=4$. The run time overhead associated with using $n_t=16$ against $n_t=4$ is easily compensated by better solver convergence. We refer to~\cite{Mang:2017SL} for a thorough study on the effect of $n_t$ on the numerical accuracy of the reduced gradient.

\subsection{Experiment 3: Registration of Mouse Brain CLARITY Images}\label{exp_2}

This experiment aims to answer both \textbf{Q1} and \textbf{Q2} by examining the performance of our scalable registration solver on ultra-high resolution mouse brain images acquired using the CLARITY imaging technique~\cite{Tomer:2014a,clarity-web}. As opposed to the previous datasets, the dataset in this experiment does not provide any real metrics for its assessment other than the relative residual (nor are we aware of any segmentation software that would work on these data).

\paragraph{Dataset} We use the \textbf{CLARITY} dataset (see~\secref{s:datasets}) for this experiment.

\paragraph{Procedure.} \textit{Preprocessing:} For all unprocessed images, the background intensity is non-zero. We normalize the image intensities such that they lie in the range [0,1] with the background intensity re-scaled to zero. Next, we affinely register all images to \texttt{Control182} at $8\times$ downsampled resolution using the SyN tool in ANTs. We report the parameter settings for the affine registration in the appendix. Subsequently, we zero-pad the images to ensure that periodic boundary conditions are satisfied for \claire{}. After preprocessing, the base image resolution is $\vect{n}_x=\vect{n}$ where $\vect{n}=(2816,3016,1162)$ and $\vect{n}/8=(328,412,1162$), respectively. We only conduct the parameter search for a single pair of images (at both resolutions independently) for these sets of images and then perform the parameter continuation on the entire dataset. We only report wall clock times for the parameter continuation and not for the parameter search.

\textit{Deformable Registration:} We register all images to the reference image \texttt{Control182} using \claire{}. We use the proposed parameter continuation scheme. We set $J_\text{min}$ to $0.05$. We do this for both resolution levels. To compare the registration accuracy between each resolution level, we follow the same steps from~\secref{exp_1b}. We compare the registration performance using the relative residual $r$. We do not have access to image segmentation for this dataset and, therefore, we cannot evaluate accuracy using Dice scores.

\begin{figure}
\centering
\includegraphics[width=\linewidth]{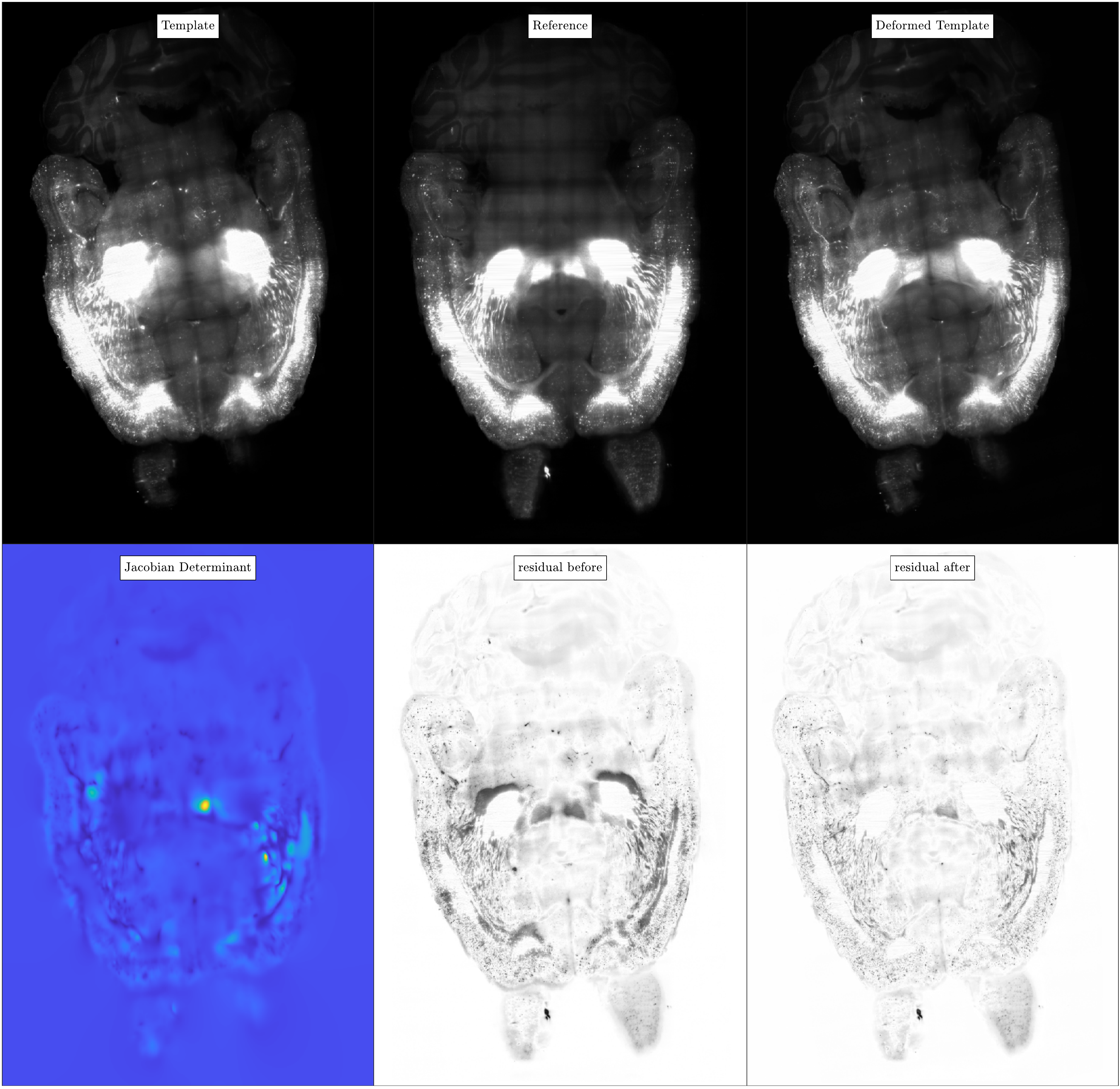}
\caption{\textbf{Experiment 3: Illustration of the registration performance for \claire{} for the CLARITY mouse brain imaging data.} We report registration results for the \texttt{Cocaine178} dataset registered to the \texttt{Control182} dataset. In row 1 (from left to right), we have the template image $m_0$ (\texttt{Cocaine178}), the reference image $m_1$ (\texttt{Control182}) and the deformed template image. The resolution of the images is $\vect{n}=(2816,3015,1162)$. In row 2, we show the determinant of the deformation gradient and the image residuals before and after registration.}\label{fig:exp_2}
\end{figure}

\begin{table}
\centering
\caption{\textbf{Experiment 3: Registration performance for \claire{} for the CLARITY imaging data at resolutions $\vect{n}=(2816,3016,1162)$ and $\vect{n}/8=(328,412,1162)$}. \texttt{Control182} is the fixed (reference) image. All other images selected from the CLARITY dataset are registered to \texttt{Control182} using a parameter continuation scheme. We fix the tolerance for the relative gradient to $5\text{e-}02$. We use linear interpolation for the semi Langrangian scheme. The bounds on the determinant $J$ of the deformation gradient for the parameter search are $[0.05,20]$. We report the estimated regularization parameters $\beta_v^\star$ and $\beta_w^\star$, the minimum and maximum values for the determinant of the deformation gradient ($J_\text{min}$ and $J_\text{max}$), the relative residual ($r$), as well as the wall clock time for the parameter continuation.}\label{tab:clarity_results_high_res}
\begin{tabular}{lllllrrrrRr}
\toprule
run &  image & \#GPU & $\vect{n}_x$ & $n_t$ & $\beta^\star_v$ & $\beta^\star_w$ & $J_{\text{min}}$ & $J_{\text{max}}$ & $r$ & runtime(s) \\
\midrule
\#1 & \multirow{2}{*}{\texttt{Fear197}} & 256 & $\vect{n}$ & 16 &  1.1e-02 & 1.0e-05 & 5.5e-02 & 2.2e+01  & 3.4e-01 & 1.4e+03 \\
\#2 & & 8 & $\vect{n}$/8 & 16 &  1.0e-03 & 1.0e-05 & 5.8e-02 & 1.5e+01  & 6.3e-01 & 9.6e+01 \\ \cline{2-11}
\#3 & \multirow{2}{*}{\texttt{Cocain178}} & 256 & $\vect{n}$ & 16 &  1.1e-02 & 1.0e-05 & 3.1e-02 & 1.2e+01  & 4.1e-01 & 6.2e+03 \\
\#4 & & 8 & $\vect{n}$/8 & 16 &  5.6e-03 & 1.0e-05 & 3.5e-02 & 4.7e+00  & 6.8e-01 & 6.1e+01 \\
\bottomrule
\end{tabular}
\end{table}

\paragraph{Results} We report the quantitative results for the registration of the CLARITY data in~\tabref{tab:clarity_results_high_res}. We showcase exemplary registration results in~\figref{fig:exp_2}.

\paragraph{Observations} The most important observation is that we can register high resolution real medical images reasonably well in under 2~hrs (see run \#1 and \#3 in~\tabref{tab:clarity_results_high_res}). Unlike the previous experiments in~\secref{exp_1a} and~\secref{exp_1b}, the reported wall clock time in~\tabref{tab:clarity_results_high_res} is for performing the parameter continuation and not the parameter search. The average time spent for the regularization parameter search for resolution $\vect{n}_x=\vect{n}$ is $\sim$2~hrs. Another observation, which is in agreement with the results reported for the experiments carried out in~\secref{exp_1a} and~\secref{exp_1b}, is that the registration performed at downsampled resolution (see~\tabref{tab:clarity_results_high_res}) results in a larger relative residual and, therefore, worse registration accuracy. We had a maximum of 256 GPUs (64 nodes, 4 GPUs per node) available to us at the TACC Longhorn supercomputer. Because of this resource constraint, our solver ran out of memory for certain parameter configurations (for example, for run \#1 and \#3, we could not use $n_t > 16$ time steps). Moreover, for all the runs in~\tabref{tab:clarity_results_high_res} we used the zero velocity approximation of $\mat{H} $ as the preconditioner and applied it at full resolution. We did not use the two-level coarse grid approximation to apply the preconditioner because it requires additional memory for the coarse grid spectral operations.

\section{Conclusions}\label{s:conclusions}

In this publication, we apply our previously developed multi-node, multi-GPU 3D image registration solver~\cite{Brunn:2020b} to study and analyze large-scale image registration. This work builds upon our former contributions on constrained large deformation diffeomorphic image registration~\cite{Mang:2015NK,Mang:2018CLAIRE,Mang:2016SC,Mang:2017SL,claire-web}. The main observations are: \emph{(i)} We are able to register CLARITY mouse brain images of unprecedented ultra-high spatial resolution ($2816\times3016\times1162$) in 23\,minutes using parameter continuation. To the best of our knowledge, images of this scale have not been registered in previous work~\cite{Brunn:2020b,Kutten:2016a,Kutten:2017b}. \emph{(ii)} We conduct detailed experiments to compare image registration performance at full and downsampled resolutions using synthetic and real images. We find that image registration at higher (native) image resolution is more accurate. To quantify the accuracy, we use Dice coefficients wherever image segmentation is available and relative residuals otherwise. We also do a sensitivity analysis for the overall solver accuracy with respect to the number of time steps $n_t$ in the SL scheme. Overall, \claire{} performed as expected: fully automatic parameter tuning works well, and higher image resolutions result in improved image similarity compared to the registration results in lower resolution. We note that these improvements in registration accuracy are less pronounced for real imaging data compared to synthetic data for the experiments conducted in this study. We attribute these observations to uncertainties and errors introduced during the registration and segmentation steps due to noise and low contrast. We discuss this in more detail in \secref{s:results}.

\paragraph{Acknowledgements and Funding.} This work was partly supported by the National Science Foundation (DMS-1854853, DMS-2012825, CCF-1817048, CCF-1725743), by the NVIDIA Corporation (NVIDIA GPU Grant Program), by the Deutsche Forschungsgemeinschaft (DFG, German Research Foundation) under Germany's Excellence Strategy-EXC 2075-390740016, by the U.S. Department of Energy, Office of Science, Office of Advanced Scientific Computing Research, Applied Mathematics program under Award Number DE-SC0019393, by the U.S. Air Force Office of Scientific Research award FA9550-17-1-0190, by the Portugal Foundation for Science and Technology and the UT Austin-Portugal program, and by NIH award 5R01NS042645-11A1. Any opinions, findings, and conclusions or recommendations expressed herein are those of the authors and do not necessarily reflect the views of the DFG, AFOSR, DOE, NIH, and NSF. Computing time on the Texas Advanced Computing Center's (TACC) systems was provided by an allocation from TACC and the NSF. This work was completed in part with resources provided by the Research Computing Data Core at the University of Houston.

\begin{appendix}
\section{Deformable registration parameters for ANTs}\label{s:ants_affine_params}
We report the deformable registration parameters for ANTs which were used for comparison with \claire{} in the parameter search experiment~\secref{s:param_cont_results}.
{\tiny
\begin{lstlisting}[language=bash,caption={ANTs registration script}]
	#!/bin/bash
	antsRegistration --dimensionality 3
	--float 1
	--output [$output_directory/,$output_directory/deformed-template.nii.gz]
	--interpolation Linear
	--winsorize-image-intensities [0.005,0.995]
	--use-histogram-matching 1
	--initial-moving-transform [$moving_image,$template_image,1]
	--transform Rigid[0.1]
	--metric MI[$reference_image,$template_image,1,32,Regular,0.25]
	--convergence [1000x500x250x100,1e-6,10]
	--shrink-factors 8x4x2x1
	--smoothing-sigmas 3x2x1x0vox
	--transform Affine[0.1]
	--metric MI[$reference_image,$template_image,1,32,Regular,0.25]
	--convergence [1000x500x250x100,1e-6,10]
	--shrink-factors 8x4x2x1
	--smoothing-sigmas 3x2x1x0vox
	--transform SyN[0.1,3,0]
	--metric MeanSquares[$reference_image,$template_image,1]
	--convergence [100x70x50x20,1e-6,10]
	--shrink-factors 8x4x2x1
	--smoothing-sigmas 3x2x1x0vox
\end{lstlisting}
}

\section{Determining $\beta_{w,\text{max}}$}\label{a:determine_betaw}

We report the runs for comparison of runtime and Dice scores for different values of $\beta_{w,\text{max}}$ for the experiment conducted in~\secref{exp_1b}.

\begin{table}[h]
\caption{\textbf{Experiment 1b: effect of $\beta_{w,\text{max}}$ on registration performance for real brain images}: Comparison of registration accuracy using Dice and relative residual $r$ for different values of $\beta_{w,\text{max}}$ at different resolutions for registration of \textbf{MRI250} brain image to \texttt{na01} and \texttt{na02} from \textbf{NIREP} dataset. We fix $\beta_{w,\text{min}}=1\text{e-}09$. We consider $\vect{n}_x=\{\vect{n}, \vect{n}/4\}$ where $\vect{n} = (640,880,880)$. We fix the tolerance for the relative gradient to $5\text{e-}02$. We use linear interpolation. The Jacobian bounds for parameter search are $[0.05,20]$. We increase the number of time steps $n_t$ proportionately with increase in resolution. We report $\beta^\star_v$ and $\beta^\star_w$, the regularization parameters from the parameter search scheme, $J_\text{min}$ and $J_\text{max}$, the minimum and maximum Jacobian determinant the relative residual $r$, average Dice $D_a$ pre and post the registration and the wall clock time for the parameter search for the registration. We highlight the best Dice scores for each resolution and for each NIREP image.}\label{tab:brainMRI250um_nirep_velocity_multilevel_Dice_stats_iporder1_betaw_variation}
\centering\tiny
\begin{tabular}{llllrrrrrrrr}
\toprule
run & NIREP & $\beta_{w,\text{max}}$ & $\vect{n}_x$ & $\beta^\star_v$ & $\beta^\star_w$ & $J_\text{min}$ & $J_\text{max}$ & $r$ & \multicolumn{2}{c}{$D_a$} & runtime(s) \\
 & & & & & & & & & pre & post & search \\
\midrule
\textbf{\#1} & \multirow{8}{*}{na01} & \multirow{2}{*}{1e-04} & $\vect{n}$ & 1.1e-05 & 1.0e-09 & 2.8e-01 & 2.7e+00 & 3.4e-01 & \multirow{8}{*}{5.5e-01} & 8.6e-01 & 5.0e+03 \\
\textbf{\#2} & & & $\vect{n}$/4 & 2.3e-05 & 1.0e-05 & 3.5e-01 & 2.8e+00 & 4.6e-01 & & 7.9e-01 & 3.5e+01 \\
\cline{3-9}\cline{11-12}
\textbf{\#3} & & \multirow{2}{*}{1e-05} & $\vect{n}$ & 1.1e-03 & 1.0e-05 & 1.8e-01 & 8.3e+00 & 2.5e-01 & & \cellcolor{light-gray}9.0e-01 & 3.1e+02 \\
\textbf{\#4} & & & $\vect{n}$/4 & 1.1e-05 & 1.0e-05 & 2.3e-01 & 8.2e+00 & 4.5e-01 & & 8.0e-01 & 3.6e+01 \\
\cline{3-9}\cline{11-12}
\textbf{\#5} & & \multirow{2}{*}{1e-06} & $\vect{n}$ & 2.0e-02 & 1.0e-06 & 5.2e-02 & 1.6e+01 & 3.0e-01 & & 8.6e-01 & 2.1e+02 \\
\textbf{\#6} & & & $\vect{n}$/4 & 1.0e-03 & 1.0e-06 & 1.2e-01 & 1.6e+01 & 3.7e-01 & & \cellcolor{light-gray}8.5e-01 & 1.2e+01 \\
\cline{3-9}\cline{11-12}
\textbf{\#7} & & \multirow{2}{*}{1e-07} & $\vect{n}$ & 5.1e-02 & 1.0e-09 & 1.8e-01 & 1.0e+01 & 4.1e-01 & & 7.9e-01 & 2.1e+02 \\
\textbf{\#8} & & & $\vect{n}$/4 & 6.6e-03 & 1.0e-07 & 1.1e-01 & 1.8e+01 & 4.0e-01 & & 8.2e-01 & 7.7e+00 \\
\cline{1-12}
\cline{3-9}\cline{11-12}
\textbf{\#9} & \multirow{8}{*}{na02} & \multirow{2}{*}{1e-04} & $\vect{n}$ & 1.1e-05 & 1.0e-09 & 2.1e-01 & 2.7e+00 & 3.4e-01 & \multirow{8}{*}{5.4e-01} & 8.4e-01 & 5.7e+03 \\
\textbf{\#10} & & & $\vect{n}$/4 & 1.1e-05 & 1.0e-06 & 8.7e-02 & 9.9e+00 & 4.7e-01 & & 7.7e-01 & 4.0e+01 \\
\cline{3-9}\cline{11-12}
\textbf{\#11} & & \multirow{2}{*}{1e-05} & $\vect{n}$ & 1.1e-03 & 1.0e-05 & 7.7e-02 & 6.3e+00 & 2.5e-01 & & \cellcolor{light-gray}8.9e-01 & 3.4e+02 \\
\textbf{\#12} & & & $\vect{n}$/4 & 1.1e-05 & 1.0e-06 & 8.1e-02 & 1.1e+01 & 4.6e-01 & & 7.7e-01 & 3.9e+01 \\
\cline{3-9}\cline{11-12}
\textbf{\#13} & & \multirow{2}{*}{1e-06} & $\vect{n}$ & 4.1e-02 & 1.0e-07 & 1.4e-01 & 2.0e+01 & 3.8e-01 &  & 8.0e-01 & 2.4e+02 \\
\textbf{\#14} & & & $\vect{n}$/4 & 1.1e-04 & 1.0e-06 & 8.2e-02 & 1.2e+01 & 4.1e-01 & & 8.1e-01 & 1.4e+01 \\
\cline{3-9}\cline{11-12}
\textbf{\#15} & & \multirow{2}{*}{1e-07} & $\vect{n}$ & 4.0e-02 & 1.0e-09 & 1.2e-01 & 1.8e+01 & 3.8e-01 &  & 8.0e-01 & 2.5e+02 \\
\textbf{\#16} & & & $\vect{n}$/4 & 1.0e-03 & 1.0e-07 & 8.8e-02 & 2.0e+01 & 3.8e-01 & & \cellcolor{light-gray}8.3e-01 & 1.4e+01 \\
\bottomrule
\end{tabular}
\end{table}

\end{appendix}


\begin{thebibliography}{100}

\bibitem{mri250}
{\em Data from: T1-weighted in vivo human whole brain mri dataset with an
  ultrahigh isotropic resolution of 250~$\mu$m}.
\newblock \url{https://datadryad.org/stash/dataset/doi:10.5061/dryad.38s74}.

\bibitem{ibmspectrum-web}
{\em {IBM} {S}pectrum {MPI} (version 10.3.0)}.

\bibitem{muse_templates}
{\em {Neuromorphometrics}}.
\newblock \url{http://www.neuromorphometrics.com}.

\bibitem{Ashburner:2007a}
{\sc J.~Ashburner}, {\em A fast diffeomorphic image registration algorithm},
  NeuroImage, 38 (2007), pp.~95--113.

\bibitem{Avants:2008a}
{\sc B.~B. Avants, C.~L. Epstein, M.~Brossman, and J.~C. Gee}, {\em Symmetric
  diffeomorphic image registration with cross-correlation: {E}valuating
  automated labeling of elderly and neurodegenerative brain}, Medical Image
  Analysis, 12 (2008), pp.~26--41.

\bibitem{ants-web}
{\sc B.~B. Avants, N.~J. Tustison, and H.~J. Johnson}, {\em {ANT}s}.

\bibitem{Avants:2011a}
{\sc B.~B. Avants, N.~J. Tustison, G.~Song, P.~A. Cook, A.~Klein, and J.~C.
  Gee}, {\em A reproducible evaluation of {ANTs} similarity metric performance
  in brain image registration}, NeuroImage, 54 (2011), pp.~2033--2044.

\bibitem{Avants:2014a}
{\sc B.~B. Avants, N.~J. Tustison, M.~Stauffer, G.~Song, B.~Wu, and J.~C. Gee},
  {\em The {{Insight ToolKit}} image registration framework}, Frontiers in
  Neuroinformatics, 0 (2014).

\bibitem{petsc-web}
{\sc S.~Balay, S.~Abhyankar, M.~F. Adams, J.~Brown, P.~Brune, K.~Buschelman,
  L.~Dalcin, A.~Dener, V.~Eijkhout, W.~D. Gropp, D.~Karpeyev, D.~Kaushik, M.~G.
  Knepley, D.~A. May, L.~C. McInnes, R.~T. Mills, T.~Munson, K.~Rupp, P.~Sanan,
  B.~F. Smith, S.~Zampini, H.~Zhang, and H.~Zhang}, {\em {PETSc} and {TAO}
  webpage ({PETSc} version 3.12.4)}.

\bibitem{petsc-efficient}
{\sc S.~Balay, W.~D. Gropp, L.~C. McInnes, and B.~F. Smith}, {\em Efficient
  management of parallelism in object oriented numerical software libraries},
  in Modern Software Tools in Scientific Computing, E.~Arge, A.~M. Bruaset, and
  H.~P. Langtangen, eds., Birkh{\"{a}}user Press, 1997, pp.~163--202.

\bibitem{Beg:2005a}
{\sc M.~F. Beg, M.~I. Miller, A.~Trouv\'e, and L.~Younes}, {\em Computing large
  deformation metric mappings via geodesic flows of diffeomorphisms},
  International Journal of Computer Vision, 61 (2005), pp.~139--157.

\bibitem{Bone:2018a}
{\sc A.~Bone, O.~Colliot, and S.~Durrleman}, {\em Learning distributions of
  shape trajectories from longitudinal datasets: {A} hierarchical model on a
  manifold of diffeomorphisms}, arXiv e-prints,  (2019).

\bibitem{Bone:2018b}
{\sc A.~Bone, M.~Louis, B.~Martin, and S.~Durrleman}, {\em Deformetrica 4: {A}n
  open-source software for statistical shape analysis}, in Proc International
  Workshop on Shape in Medical Imaging, vol.~LNCS 11167, 2018, pp.~3--13.

\bibitem{Brunn:2020b}
{\sc M.~Brunn, N.~Himthani, G.~Biros, M.~Mehl, and A.~Mang}, {\em Multi-{{Node
  Multi}}-{{GPU Diffeomorphic Image Registration}} for {{Large}}-{{Scale
  Imaging Problems}}}, in {{SC20}}: International {{Conference}} for {{High
  Performance Computing}}, {{Networking}}, {{Storage}} and {{Analysis}}, Nov.
  2020, pp.~1--17.

\bibitem{Brunn2021}
{\sc M.~Brunn, N.~Himthani, G.~Biros, M.~Mehl, and A.~Mang}, {\em Claire:
  Constrained large deformation diffeomorphic image registration on parallel
  computing architectures}, Journal of Open Source Software, 6 (2021), p.~3038.

\bibitem{Brunn:2020a}
{\sc M.~Brunn, N.~Himthani, G.~Biros, M.~Mehl, and A.~Mang}, {\em Fast {{GPU
  3D}} diffeomorphic image registration}, Journal of Parallel and Distributed
  Computing, 149 (2021), pp.~149--162.

\bibitem{Budelmann:2019a}
{\sc D.~Budelmann, L.~Koenig, N.~Papenberg, and J.~Lellmann}, {\em
  Fully-deformable {3D} image registration in two seconds}, in Bildverarbeitung
  f\"ur die Medizin, 2019, pp.~302--307.

\bibitem{lddmm-web}
{\sc {Center for Imaging Science, Johns Hopkins University}}, {\em {LDDMM}
  {S}uite}.

\bibitem{clarity-web}
{\sc V.~Chandrashekhar, A.~Crow, J.~Bogelstein, and K.~Deisseroth}, {\em
  {NEURODATA CLARITOMES}}.

\bibitem{Christensen:2006a}
{\sc G.~E. Christensen, X.~Geng, J.~G. Kuhl, J.~Bruss, T.~J. Grabowski, I.~A.
  Pirwani, M.~W. Vannier, J.~S. Allen, and H.~Damasio}, {\em Introduction to
  the non-rigid image registration evaluation project}, in Proc Biomedical
  Image Registration, vol.~LNCS 4057, 2006, pp.~128--135.

\bibitem{Chung:2013a}
{\sc K.~Chung and K.~Deisseroth}, {\em Clarity for mapping the nervous system},
  Nature methods, 10 (2013), pp.~508--513.

\bibitem{Chung:2013b}
{\sc K.~Chung, J.~Wallace, S.-Y. Kim, S.~Kalyanasundaram, A.~S. Andalman, T.~J.
  Davidson, J.~J. Mirzabekov, K.~A. Zalocusky, J.~Mattis, A.~K. Denisin,
  S.~Pak, H.~Bernstein, L.~G.~C. Ramakrishnan, V.~Gradinaru, and
  K.~Deisseroth}, {\em Structural and molecular interrogation of intact
  biological systems}, Nature, 497 (2013), pp.~332--337.

\bibitem{Courty:2008a}
{\sc N.~Courty and P.~Hellier}, {\em Accelerating {3D} non-rigid registration
  using graphics hardware}, International Journal of Image and Graphics, 8
  (2008), pp.~81--98.

\bibitem{Doshi:2016a}
{\sc J.~Doshi, G.~Erus, Y.~Ou, S.~M. Resnick, R.~C. Gur, R.~E. Gur, T.~D.
  Satterthwaite, S.~Furth, and C.~Davatzikos}, {\em {{MUSE}}: {{MUlti-atlas}}
  region {{Segmentation}} utilizing {{Ensembles}} of registration algorithms
  and parameters, and locally optimal atlas selection}, NeuroImage, 127 (2016),
  pp.~186--195.

\bibitem{deformetrica-web}
{\sc S.~Durrleman, A.~Brone, M.~Louis, B.~Martin, P.~Gori, A.~Routier,
  M.~Bacci, A.~Fouquier, B.~Charlier, J.~Glaunes, J.~Fishbaugh, M.~Prastawa,
  M.~Diaz, and C.~Doucet}, {\em deformetrica}.

\bibitem{Durrleman:2014a}
{\sc S.~Durrleman, M.~Prastawa, N.~Charon, J.~R. Korenberg, S.~Joshi, G.~Gerig,
  and A.~Trouve}, {\em Morphometry of anatomical shape complexes with dense
  deformations and sparse parameters}, NeuroImage, 101 (2014), pp.~35--49.

\bibitem{Eklund:2013a}
{\sc A.~Eklund, P.~Dufort, D.~Forsberg, and S.~M. LaConte}, {\em Medical image
  processing on the {GPU}--past, present and future}, Medical Image Analysis,
  17 (2013), pp.~1073--1094.

\bibitem{Ellingwood:2016a}
{\sc N.~D. Ellingwood, Y.~Yin, M.~Smith, and C.-L. Lin}, {\em Efficient methods
  for implementation of multi-level nonrigid mass-preserving image registration
  on {GPU}s and multi-threaded {CPU}s}, Computer Methods and Programs in
  Biomedicine, 127 (2016), pp.~290--300.

\bibitem{Fischer:2008a}
{\sc B.~Fischer and J.~Modersitzki}, {\em Ill-posed medicine -- an introduction
  to image registration}, Inverse Problems, 24 (2008), pp.~1--16.

\bibitem{Fishbaugh:2017a}
{\sc J.~Fishbaugh, S.~Durrleman, M.~Prastawa, and G.~Gerig}, {\em Geodesic
  shape regression with multiple geometries and sparse parameters}, Medical
  Image Analysis, 39 (2017), pp.~1--17.

\bibitem{niftilib-web}
{\sc K.~Fissell and R.~Reynolds}, {\em niftilib (version 2.2.0)}, 2020.

\bibitem{Fluck:2011a}
{\sc O.~Fluck, C.~Vetter, W.~Wein, A.~Kamen, B.~Preim, and R.~Westermann}, {\em
  A survey of medical image registration on graphics hardware}, Computer
  Methods and Programs in Biomedicine, 104 (2011), pp.~e45--e57.

\bibitem{accfft_github}
{\sc A.~Gholami and G.~Biros}, {\em {AccFFT}}, 2017.

\bibitem{accfft-home-page}
{\sc A.~Gholami and G.~Biros}, {\em {AccFFT} home page}, 2017.

\bibitem{Gholami:2016IP}
{\sc A.~Gholami, A.~Mang, and G.~Biros}, {\em An inverse problem formulation
  for parameter estimation of a reaction-diffusion model of low grade gliomas},
  Journal of Mathematical Biology, 72 (2016), pp.~409--433.

\bibitem{Gholami:2017SC}
{\sc A.~Gholami, A.~Mang, K.~Scheufele, C.~Davatzikos, M.~Mehl, and G.~Biros},
  {\em A framework for scalable biophysics-based image analysis}, in Proc
  ACM/IEEE Conference on Supercomputing, 2017, pp.~1--13.

\bibitem{Grzech:2019a}
{\sc D.~Grzech, L.~Folgoc, M.~P. Heinrich, B.~Khanal, J.~Moll, J.~A. Schnabel,
  B.~Glocker, and B.~Kainz}, {\em {FastReg}: {F}ast non-rigid registration via
  accelerated optimisation on the manifold of diffeomorphisms}, arXiv e-prints,
   (2019).

\bibitem{Gu:2009a}
{\sc X.~Gu, H.~Pan, Y.~Liang, R.~Castillo, D.~Yang, D.~Choi, E.~Castillo,
  A.~Majumdar, T.~Guerrero, and S.~B. Jiang}, {\em Implementation and
  evaluation of various demons deformable image registration algorithms on a
  {GPU}}, Physics in Medicine and Biology, 55 (2009), pp.~207--219.

\bibitem{Ha:2011a}
{\sc L.~Ha, J.~Kr\"uger, S.~Joshi, and C.~T. Silva}, {\em Multiscale unbiased
  diffeomorphic atlas construction on multi-{GPU}s}, in CPU Computing Gems
  Emerald Edition, Elsevier Inc, 2011, ch.~48, pp.~771--791.

\bibitem{Ha:2009a}
{\sc L.~K. Ha, J.~Kr\"uger, P.~T. Fletcher, S.~Joshi, and C.~T. Silva}, {\em
  Fast parallel unbiased diffeomorphic atlas construction on multi-graphics
  processing units}, in Proc Eurographics Conference on Parallel Grphics and
  Visualization, 2009, pp.~41--48.

\bibitem{haberGCVBasedMethod2000}
{\sc E.~Haber and D.~Oldenburg}, {\em A {{GCV}} based method for nonlinear
  ill-posed problems}, Computational Geosciences, 4 (2000), pp.~41--63.

\bibitem{Hajnal:2001a}
{\sc J.~V. Hajnal, D.~L.~G. Hill, and D.~J. Hawkes}, eds., {\em Medical Image
  Registration}, CRC Press, Boca Raton, Florida, US, 2001.

\bibitem{Hernandez:2009a}
{\sc M.~Hernandez, M.~N. Bossa, and S.~Olmos}, {\em Registration of anatomical
  images using paths of diffeomorphisms parameterized with stationary vector
  field flows}, International Journal of Computer Vision, 85 (2009),
  pp.~291--306.

\bibitem{Thrust}
{\sc J.~Hoberock and N.~Bell}, {\em Thrust, the cuda c++ template library},
  2010.

\bibitem{ibmxl-web}
{\sc {IBM}}, {\em {IBM XL C/C++} (version 16.1.1)}.

\bibitem{itkndreg-web}
{\sc {Insight Software Consortium}}, {\em {ITKNDReg}}.

\bibitem{jenkinson2012fsl}
{\sc M.~Jenkinson, C.~F. Beckmann, T.~E. Behrens, M.~W. Woolrich, and S.~M.
  Smith}, {\em Fsl}, Neuroimage, 62 (2012), pp.~782--790.

\bibitem{jenkinsonFSL2012a}
{\sc M.~Jenkinson, C.~F. Beckmann, T.~E.~J. Behrens, M.~W. Woolrich, and S.~M.
  Smith}, {\em {{FSL}}}, NeuroImage, 62 (2012), pp.~782--790.

\bibitem{Joshi:2005a}
{\sc S.~Joshi, B.~Davis, M.~Jornier, and G.~Gerig}, {\em Unbiased diffeomorphic
  atlas construction for computational anatomy}, NeuroImage, 23 (2005),
  pp.~S151--S160.

\bibitem{Kim:2013a}
{\sc S.-Y. Kim, K.~Chung, and K.~Deisseroth}, {\em Light microscopy mapping of
  connections in the intact brain}, Trends in Cognitive Sciences, 17 (2013),
  pp.~596--599.

\bibitem{Klein:2010a}
{\sc S.~Klein, M.~Staring, K.~Murphy, M.~A. Viergever, and J.~P.~W. Pluim},
  {\em {ELASTIX}: {A} tollbox for intensity-based medical image registration},
  Medical Imaging, IEEE Transactions on, 29 (2010), pp.~196--205.

\bibitem{Koenig:2018a}
{\sc L.~Koenig, J.~Ruehaak, A.~Derksen, and J.~Lellmann}, {\em A matrix-free
  approach to parallel and memory-efficient deformable image registration},
  SIAM Journal on Scientific Computing, 40 (2018), pp.~B858--B888.

\bibitem{Krebs:2018a}
{\sc J.~Krebs, T.~Mansi, B.~Mailh\'e, N.~Ayache, and H.~Delingette}, {\em
  Unsupervised probabilistic deformation modeling for robust diffeomorphic
  registration}, in Proc International Workshop on Deep Learning in Medical
  Image Analysis, vol.~LNCS 11045, 2018, pp.~101--109.

\bibitem{Kuan:2015a}
{\sc L.~Kuan, Y.~Li, C.~Lau, D.~Feng, A.~Bernard, S.~M. Sunkin, H.~Zeng,
  C.~Dang, M.~Hawrylycz, and L.~Ng}, {\em Neuroinformatics of the {{Allen Mouse
  Brain Connectivity Atlas}}}, Methods, 73 (2015), pp.~4--17.

\bibitem{Kutten:2016a}
{\sc K.~S. Kutten, N.~Charon, M.~I. Miller, J.~T. Ratnanather, K.~Deisseroth,
  L.~Ye, and J.~T. Vogelstein}, {\em A diffeomorphic approach to multimodal
  registration with mutual information: {A}pplications to {CLARITY} mouse brain
  images}, ArXiv e-prints,  (2016).

\bibitem{Kutten:2017a}
\leavevmode\vrule height 2pt depth -1.6pt width 23pt, {\em A diffeomorphic
  approach to multimodal registration with mutual information: {A}pplications
  to {CLARITY} mouse brain images}, in Proc Medical Image Computing and
  Computer-Assisted Intervention, vol.~LNCS 10433, 2017, pp.~275--282.

\bibitem{Kutten:2017b}
{\sc K.~S. Kutten, N.~Charon, M.~I. Miller, J.~T. Ratnanather, J.~Matelsky,
  A.~D. Baden, K.~Lillaney, K.~Deisseroth, L.~Ye, and J.~T. Vogelstein}, {\em A
  {{Large Deformation Diffeomorphic Approach}} to {{Registration}} of {{CLARITY
  Images}} via {{Mutual Information}}}, arXiv:1612.00356 [cs],  (2017).

\bibitem{Kutten:2016a}
{\sc K.~S. Kutten, J.~T. Vogelstein, N.~Charon, L.~Ye, K.~D. M.d, and M.~I.
  Miller}, {\em Deformably registering and annotating whole {{CLARITY}} brains
  to an atlas via masked {{LDDMM}}}, in Optics, {{Photonics}} and {{Digital
  Technologies}} for {{Imaging Applications IV}}, vol.~9896, {SPIE}, Apr. 2016,
  pp.~282--290.

\bibitem{pnetcdflib}
{\sc R.~Latham, M.~Zingale, R.~Thakur, W.~Gropp, B.~Gallagher, W.~Liao,
  A.~Siegel, R.~Ross, A.~Choudhary, and J.~Li}, {\em Parallel netcdf: A
  high-performance scientific i/o interface}, in SC Conference, Los Alamitos,
  CA, USA, nov 2003, IEEE Computer Society, p.~39.

\bibitem{Lorenzi:2013a}
{\sc M.~Lorenzi and X.~Pennec}, {\em Geodesics, parallel transport and
  one-parameter subgroups for diffeomorphic image registration}, International
  Journal of Computer Vision, 105 (2013), pp.~111--127.

\bibitem{Lusebrink:2017a}
{\sc F.~L{\"u}sebrink, A.~Sciarra, H.~Mattern, R.~Yakupov, and O.~Speck}, {\em
  T1-weighted in vivo human whole brain {{MRI}} dataset with an ultrahigh
  isotropic resolution of 250 {$\mu$}m}, Scientific Data, 4 (2017), p.~170032.

\bibitem{Mang:2020a}
{\sc A.~Mang, S.~Bakas, S.~Subramanian, C.~Davatzikos, and G.~Biros}, {\em
  Integrated biophysical modeling and image analysis: {A}pplication to
  neuro-oncology}, Annual Review of Biomedical Engineering, 22 (2020),
  pp.~309--341.

\bibitem{Mang:2015NK}
{\sc A.~Mang and G.~Biros}, {\em An inexact {N}ewton--{K}rylov algorithm for
  constrained diffeomorphic image registration}, SIAM Journal on Imaging
  Sciences, 8 (2015), pp.~1030--1069.

\bibitem{Mang:2016H1}
\leavevmode\vrule height 2pt depth -1.6pt width 23pt, {\em Constrained
  {$H^1$}-regularization schemes for diffeomorphic image registration}, SIAM
  Journal on Imaging Sciences, 9 (2016), pp.~1154--1194.

\bibitem{Mang:2017SL}
\leavevmode\vrule height 2pt depth -1.6pt width 23pt, {\em A
  {S}emi-{L}agrangian two-level preconditioned {N}ewton--{K}rylov solver for
  constrained diffeomorphic image registration}, SIAM Journal on Scientific
  Computing, 39 (2017), pp.~B1064--B1101.

\bibitem{claire-web}
{\sc A.~Mang and G.~Biros}, {\em Constrained large deformation diffeomorphic
  image registration ({CLAIRE})}, 2019.
\newblock [Commit: v0.07-131-gbb7619e].

\bibitem{Mang:2016SC}
{\sc A.~Mang, A.~Gholami, and G.~Biros}, {\em Distributed-memory
  large-deformation diffeomorphic {3D} image registration}, in Proc ACM/IEEE
  Conference on Supercomputing, 2016.

\bibitem{Mang:2018a}
{\sc A.~Mang, A.~Gholami, C.~Davatzikos, and G.~Biros}, {\em {PDE}-constrained
  optimization in medical image analysis}, Optimization and Engineering, 19
  (2018), pp.~765--812.
\newblock \url{https://doi.org/10.1007/s11081-018-9390-9}.

\bibitem{Mang:2018CLAIRE}
{\sc A.~Mang, A.~Gholami, C.~Davatzikos, and G.~Biros}, {\em {CLAIRE:} a
  distributed-memory solver for constrained large deformation diffeomorphic
  image registration}, SIAM Journal on Scientific Computing, 41 (2019),
  pp.~C548--C584.

\bibitem{Mang:2017a}
{\sc A.~Mang and L.~Ruthotto}, {\em A {L}agrangian {G}auss--{N}ewton--{K}rylov
  solver for mass- and intensity-preserving diffeomorphic image registration},
  SIAM Journal on Scientific Computing, 39 (2017), pp.~B860--B885.

\bibitem{Modat:2010a}
{\sc M.~Modat, G.~R. Ridgway, Z.~A. Taylor, M.~Lehmann, J.~Barnes, D.~J.
  Hawkes, N.~C. Fox, and S.~Ourselin}, {\em Fast free-form deformation using
  graphics processing units}, Computer Methods and Programs in Biomedicine, 98
  (2010), pp.~278--284.

\bibitem{Modersitzki:2004a}
{\sc J.~Modersitzki}, {\em Numerical methods for image registration}, Oxford
  University Press, New York, 2004.

\bibitem{Modersitzki:2008a}
\leavevmode\vrule height 2pt depth -1.6pt width 23pt, {\em {FLIRT} with
  rigidity---image registration with a local non-rigidity penalty},
  International Journal of Computer Vision, 76 (2008), pp.~153--163.

\bibitem{Nazib:2018a}
{\sc A.~Nazib, J.~Galloway, C.~Fookes, and D.~Perrin}, {\em Performance of
  {{Registration Tools}} on {{High-Resolution 3D Brain Images}}}, in 2018 40th
  {{Annual International Conference}} of the {{IEEE Engineering}} in
  {{Medicine}} and {{Biology Society}} ({{EMBC}}), July 2018, pp.~566--569.

\bibitem{ardent-web}
{\sc neurodata}, {\em {ARDENT}}.

\bibitem{Niedworok:2016a}
{\sc C.~J. Niedworok, A.~P.~Y. Brown, M.~Jorge~Cardoso, P.~Osten, S.~Ourselin,
  M.~Modat, and T.~W. Margrie}, {\em {{aMAP}} is a validated pipeline for
  registration and segmentation of high-resolution mouse brain data}, Nature
  Communications, 7 (2016), p.~11879.

\bibitem{cuda-web}
{\sc {NVIDIA}}, {\em {CUDA Toolkit} (version 10.1)}.

\bibitem{Nvidia2007b}
{\sc Nvidia}, {\em {CUDA CUFFT Library}}, 2007.

\bibitem{pyca-web}
{\sc J.~S. Preston}, {\em Python for computational anatomy}.

\bibitem{Scheufele:2019a}
{\sc K.~Scheufele, A.~Mang, A.~Gholami, C.~Davatzikos, G.~Biros, and M.~Mehl},
  {\em Coupling brain-tumor biophysical models and diffeomorphic image
  registration}, Computer Methods in Applied Mechanics and Engineering, 347
  (2019), pp.~533--567.

\bibitem{Scheufele:2020a}
{\sc K.~Scheufele, S.~Subramanian, A.~Mang, G.~Biros, and M.~Mehl}, {\em
  Image-driven biophysical tumor growth model calibration}, SIAM Journal on
  Scientific Computing, 42 (2020), pp.~B549--B580.

\bibitem{Shackleford:2010a}
{\sc J.~Shackleford, N.~Kandasamy, and G.~Sharp}, {\em On developing {B}-spline
  registration algorithms for multi-core processors}, Physics in Medicine and
  Biology, 55 (2010), pp.~6329--6351.

\bibitem{Shamonin:2014a}
{\sc D.~P. Shamonin, E.~E. Bron, B.~P.~F. Lelieveldt, M.~Smits, S.~Klein, and
  M.~Staring}, {\em Fast parallel image registration on {CPU} and {GPU} for
  diagnostic classification of {A}lzheimer's disease}, Frontiers in
  Neuroinformatics, 7 (2014), pp.~1--15.

\bibitem{Shams:2010a}
{\sc R.~Shams, P.~Sadeghi, R.~A. Kennedy, and R.~I. Hartley}, {\em A survey of
  medical image registration on multicore and the {GPU}}, Signal Processing
  Magazine, IEEE, 27 (2010), pp.~50--60.

\bibitem{Smith:2004a}
{\sc S.~M. Smith, M.~Jenkinson, M.~W. Woolrich, C.~F. Beckmann, T.~E. Behrens,
  H.~Johansen-Berg, P.~R. Bannister, M.~De~Luca, I.~Drobnjak, D.~E. Flitney,
  et~al.}, {\em Advances in functional and structural mr image analysis and
  implementation as fsl}, Neuroimage, 23 (2004), pp.~S208--S219.

\bibitem{Sommer:2011a}
{\sc S.~Sommer}, {\em Accelerating multi-scale flows for {LDDKBM} diffeomorphic
  registration}, in Proc IEEE International Conference on Computer Visions
  Workshops, 2011, pp.~499--505.

\bibitem{Sotiras:2013a}
{\sc A.~Sotiras, C.~Davatzikos, and N.~Paragios}, {\em Deformable medical image
  registration: {A} survey}, Medical Imaging, IEEE Transactions on, 32 (2013),
  pp.~1153--1190.

\bibitem{Susaki:2015a}
{\sc E.~A. Susaki, K.~Tainaka, D.~Perrin, H.~Yukinaga, A.~Kuno, and H.~R.
  Ueda}, {\em Advanced {{CUBIC}} protocols for whole-brain and whole-body
  clearing and imaging}, Nature Protocols, 10 (2015), pp.~1709--1727.

\bibitem{Tomer:2014a}
{\sc R.~Tomer, L.~Ye, B.~Hsueh, and K.~Deisseroth}, {\em Advanced {CLARITY} for
  rapid and high-resolution imaging of intact tissues}, Nature protocols, 9
  (2014), pp.~1682--1697.

\bibitem{Trouve:1998a}
{\sc A.~Trouv\'e}, {\em Diffeomorphism groups and pattern matching in image
  analysis}, International Journal of Computer Vision, 28 (1998), pp.~213--221.

\bibitem{ValeroLara:2013a}
{\sc P.~Valero-Lara}, {\em A {GPU} approach for accelerating {3D} deformable
  registration ({DARTEL}) on brain biomedical images}, in Proc European MPI
  Users' Group Meeting, 2013, pp.~187--192.

\bibitem{ValeroLara:2014a}
{\sc P.~Valero-Lara}, {\em Multi-{GPU} acceleration of {DARTEL} (early
  detection of {A}lzheimer)}, in Proc IEEE International Conference on Cluster
  Computing, 2014, pp.~346--354.

\bibitem{Vercauteren:2007a}
{\sc T.~Vercauteren, X.~Pennec, A.~Perchant, and N.~Ayache}, {\em Diffeomorphic
  demons using {ITK}'s finite difference solver hierarchy}, The Insight
  Journal, 1926/510 (2007).
\newblock
  \href{http://hdl.handle.net/1926/510}{http://hdl.handle.net/1926/510}.

\bibitem{Vercauteren:2009a}
\leavevmode\vrule height 2pt depth -1.6pt width 23pt, {\em Diffeomorphic
  demons: {E}fficient non-parametric image registration}, NeuroImage, 45
  (2009), pp.~S61--S72.

\bibitem{Vogelstein:2018b}
{\sc J.~T. Vogelstein, E.~Perlman, B.~Falk, A.~Baden, W.~Gray~Roncal,
  V.~Chandrashekhar, F.~Collman, S.~Seshamani, J.~L. Patsolic, K.~Lillaney,
  et~al.}, {\em A community-developed open-source computational ecosystem for
  big neuro data}, Nature methods, 15 (2018), pp.~846--847.

\bibitem{Vogelstein:2018a}
{\sc J.~T. Vogelstein, E.~Perlman, B.~Falk, A.~Baden, W.~G. Roncal,
  V.~Chandrashekhar, F.~Collman, S.~Seshamani, J.~L. Patsolic, K.~Lillaney,
  M.~Kazhdan, R.~Hider, D.~Pryor, J.~Matelsky, T.~Gion, P.~Manavalan,
  B.~Wester, M.~Chevillet, E.~T. Trautman, K.~Khairy, E.~Bridgeford, D.~M.
  Kleissas, D.~J. Tward, A.~K. Crow, B.~Hsueh, M.~A. Wright, M.~I. Miller,
  S.~J. Smith, R.~J. Vogelstein, K.~Deisseroth, and R.~Burns}, {\em A
  community-developed open-source computational ecosystem for big neuro data},
  Nature Methods, 11 (2018), pp.~846--847.

\bibitem{Woolrich:2009a}
{\sc M.~W. Woolrich, S.~Jbabdi, B.~Patenaude, M.~Chappell, S.~Makni,
  T.~Behrens, C.~Beckmann, M.~Jenkinson, and S.~M. Smith}, {\em Bayesian
  analysis of neuroimaging data in fsl}, Neuroimage, 45 (2009), pp.~S173--S186.

\bibitem{Yang:2017a}
{\sc X.~Yang, R.~Kwitt, M.~Styner, and M.~Niethammer}, {\em Quicksilver: {F}ast
  predictive image registration---{A} deep learning approach}, NeuroImage, 158
  (2017), pp.~378--396.

\bibitem{Younes:2010a}
{\sc L.~Younes}, {\em Shapes and diffeomorphisms}, Springer, 2010.

\bibitem{itksnap}
{\sc P.~A. Yushkevich, J.~Piven, H.~Cody~Hazlett, R.~Gimpel~Smith, S.~Ho, J.~C.
  Gee, and G.~Gerig}, {\em User-guided {3D} active contour segmentation of
  anatomical structures: Significantly improved efficiency and reliability},
  Neuroimage, 31 (2006), pp.~1116--1128.

\bibitem{Zhang:2018a}
{\sc M.~Zhang and P.~T. Fletcher}, {\em Fast diffeomorphic image registration
  via {F}ourier-approximated {L}ie algebras}, International Journal of Computer
  Vision,  (2018), pp.~1--13.

\bibitem{Zhang:2019a}
{\sc M.~Zhang and P.~T. Fletcher}, {\em Fast {{Diffeomorphic Image
  Registration}} via {{Fourier-Approximated Lie Algebras}}}, International
  Journal of Computer Vision, 127 (2019), pp.~61--73.

\bibitem{Zhang:2001a}
{\sc Y.~Zhang, M.~Brady, and S.~Smith}, {\em Segmentation of brain mr images
  through a hidden markov random field model and the expectation-maximization
  algorithm}, IEEE transactions on medical imaging, 20 (2001), pp.~45--57.

\end{thebibliography}
\end{document}